%% file: main.tex
\documentclass{article} 
\PassOptionsToPackage{table}{xcolor}
\usepackage[final]{colm2026_conference}

\usepackage{microtype}
\usepackage{graphicx}
\usepackage{booktabs} 

\usepackage{hyperref}

\usepackage{amsmath}
\usepackage{amssymb}
\usepackage{mathtools}
\usepackage{amsthm}
\usepackage{makecell}

\usepackage[textsize=tiny]{todonotes}

\input{macros}

\input{math_commands.tex}

\theoremstyle{remark}

\usepackage{pifont}
\usepackage{caption}

\usepackage{fontawesome}

\newcommand{\mycolorbox}[2]{\tcbox[colback=#1, colframe=myred,
       boxrule=0pt, arc=2pt, top=0pt, bottom=0pt, left=2pt, right=2pt, boxsep=0pt, on line]{#2}}
\newcommand{\llama}{Llama3 8B Instruct}

\usepackage{lineno}

\definecolor{darkblue}{rgb}{0, 0, 0.5}
\hypersetup{colorlinks=true, citecolor=darkblue, linkcolor=darkblue, urlcolor=darkblue}

\title{LLMs Can Get ``Brain Rot'': A Pilot Study on Twitter/X}

\newcommand{\utexas}{\textsuperscript{2}}
\newcommand{\tamu}{\textsuperscript{1}}
\newcommand{\purdue}{\textsuperscript{3}}

\author{Shuo Xing$^\dagger$\tamu, Junyuan Hong$^{\dagger*}$\utexas, Yifan Wang$^\ddag$\purdue, Runjin Chen$^\ddag$\utexas, Zhenyu Zhang\utexas, \\
\textbf{Ananth Grama\purdue, Zhengzhong Tu\tamu, Zhangyang Wang\utexas}\thanks{Correspondence to \texttt{jyhong@nus.edu.sg}, \texttt{atlaswang@utexas.edu}. $^\dagger$Lead authors with equal contributions. $^\ddag$Core contributors.
} \\
       \tamu Texas A\&M University, 
       \utexas University of Texas at Austin, \purdue Purdue University \\
       \faGlobe~Model \& Code: \url{https://llm-brain-rot.github.io/}
       }

\begin{document}

\ifcolmsubmission
\linenumbers
\fi

\maketitle

\begin{abstract}
  We propose and test the \textbf{LLM Brain Rot Hypothesis}: continual exposure to \textit{junk web text} induces lasting cognitive decline in large language models (LLMs). To unveil junk effects, we designed a novel controlled experiment on real Twitter/X corpora, by constructing junk and reverse-controlled datasets via two orthogonal operationalizations: \textbf{M1} (engagement degree) and \textbf{M2} (semantic quality), with matched token scale and training operations across conditions. Compared to the control group, continual pre-training of 4 LLMs on the junk dataset causes non-trivial declines (Hedges' $g>0.3$) on reasoning, long-context understanding, safety, and inflating ``dark traits'' (e.g., psychopathy, narcissism). The gradual mixtures of junk and control datasets also yield dose-response cognition decay: for example, under M1, ARC-Challenge with Chain-of-Thought drops $72.1 \rightarrow 57.2$ and RULER-CWE $83.7 \rightarrow 52.3$ as junk ratio rises from $0\%$ to $100\%$.

  Error forensics reveal several key insights. First, we identify \textit{thought-skipping as the primary lesion in reasoning}: models increasingly truncate or skip chains. Second, partial but incomplete healing is observed: scaling instruction tuning and clean continual pre-training improve the declined cognition, yet cannot restore baseline capability, suggesting persistent representational drift rather than format mismatch. Finally, we discover that the popularity, a non-semantic metric, of a tweet is a better indicator of the Brain Rot effect than the length in M1. Together, the results provide significant, multi-perspective evidence that \textit{social effects of data could be a causal driver of LLM capability decay in continual pre-training}, thereby motivating routine ``cognitive health checks'' for deployed and evolving LLMs.
\end{abstract}

\input{sec/intro}
\input{sec/related}

\input{sec/exp}
\input{sec/conclusion}


\input{sec/ack}


\bibliography{main}
\bibliographystyle{colm2026_conference}

\newpage
\appendix


\input{sec/appd/related}

\input{sec/appd/exp}

\input{sec/discussion}
\end{document}

%% file: macros.tex
\usepackage[table]{xcolor}  
\definecolor{myred}{HTML}{E14169} 
\definecolor{royalblue}{HTML}{4169E1}
\definecolor{mycyan}{HTML}{0AB6D8}
\definecolor{myblue}{HTML}{2766F5}  
\definecolor{mydeepblue}{HTML}{004E98} 
\definecolor{mygreen}{HTML}{D1FFBD}

\usepackage{url}            
\usepackage{amsthm}
\usepackage{amssymb}
\usepackage{mathtools}
\usepackage{hyperref}
\usepackage[nameinlink,capitalize]{cleveref}
\usepackage[normalem]{ulem} 

\usepackage[utf8]{inputenc} 
\usepackage[T1]{fontenc}    
\usepackage{booktabs}       
\usepackage{amsfonts}       
\usepackage{nicefrac}       
\usepackage{microtype}      
\usepackage{proof-at-the-end}  
\usepackage{multirow}
\usepackage{lscape}         

\usepackage{graphicx,wrapfig}
\usepackage{amsfonts}
\usepackage{url}
\usepackage{enumitem}
\usepackage[caption=false,font=normalsize,labelfont=sf,textfont=sf]{subfig}
\usepackage{amsthm}
\usepackage{thmtools}
\usepackage{thm-restate}

\usepackage{listings}
\definecolor{codecomment}{rgb}{0,0.6,0}   
\definecolor{codekeyword}{rgb}{0,0,1}     
\definecolor{codestring}{rgb}{0.8,0,0}    

\lstdefinestyle{mypython}{
    language=Python,
    backgroundcolor=\color{royalblue!5!white},   
    basicstyle=\ttfamily\tiny, 
    keywordstyle=\color{codekeyword}\bfseries,
    stringstyle=\color{codestring},
    commentstyle=\color{codecomment}\itshape,
    numbers=left,
    numberstyle=\tiny\color{gray},
    frame=single,                     
    rulecolor=\color{black},          
    breaklines=true,                  
    showstringspaces=false
}

\usepackage[many]{tcolorbox}
\newtcolorbox{boxK}[2][]{
    sharpish corners, 
    boxrule = 0pt,
    toprule = 4.5pt, 
    enhanced,
    fuzzy shadow = {0pt}{-2pt}{-0.5pt}{0.5pt}{black!35}, 
    fontupper = \ttfamily\small, 
    boxsep = 5pt, 
    left = 5pt, 
    right = 5pt, 
    top = 5pt, 
    bottom = 5pt, 
    #1                       
}

\newtcolorbox{takeawaybox}[2][]{
    enhanced,
    boxsep = 2pt, 
    left = 2pt, 
    right = 2pt, 
    top = 2pt, 
    bottom = 2pt, 
    #1                       
} 

\newcommand{\bc}{\begin{center}}
\newcommand{\ec}{\end{center}}

\newcommand{\bdm}{\begin{displaymath}}
\newcommand{\edm}{\end{displaymath}}

\newcommand{\beq}{\begin{equation}}
\newcommand{\eeq}{\end{equation}}

\newcommand{\bfl}{\begin{flushleft}}
\newcommand{\efl}{\end{flushleft}}

\newcommand{\bt}{\begin{tabbing}}
\newcommand{\et}{\end{tabbing}}

\newcommand{\beqn}{\begin{align}}
\newcommand{\eeqn}{\end{align}}

\newcommand{\beqs}{\begin{align*}} 
\newcommand{\eeqs}{\end{align*}}  


%% file: math_commands.tex
\usepackage{amsmath,amsfonts,bm}

\def\eqref#1{equation~\ref{#1}}

\def\1{\bm{1}}

\DeclareMathAlphabet{\mathsfit}{\encodingdefault}{\sfdefault}{m}{sl}
\SetMathAlphabet{\mathsfit}{bold}{\encodingdefault}{\sfdefault}{bx}{n}



%% file: sec/intro.tex
\section{Introduction}
\vspace{-3mm}
In 2024, the term ``Brain Rot'' was named the Oxford word of year~\citep{oxford_brain_rot} when it drew increasing concern in modern society.
Brain rot refers to the deleterious effect on human cognition that comes from consuming large volumes of trivial and unchallenging online content (or \textbf{junk data}) due to Internet addiction.
Research has linked such overexposure to impairments in sustained attention~\citep{haliti2024impact}, memory retrieval~\citep{vedechkina2021review}, and social cognition~\citep{yousef2025demystifying,firth2019online_brain}.
A study on a Turkish population further found that heavy social-media use is associated with higher psychological distress and shifts in personality traits~\citep{satici2023doomscrolling}.

\begin{figure}[ht]
    \centering
    \vspace{-0.15in}
    \includegraphics[width=\linewidth]{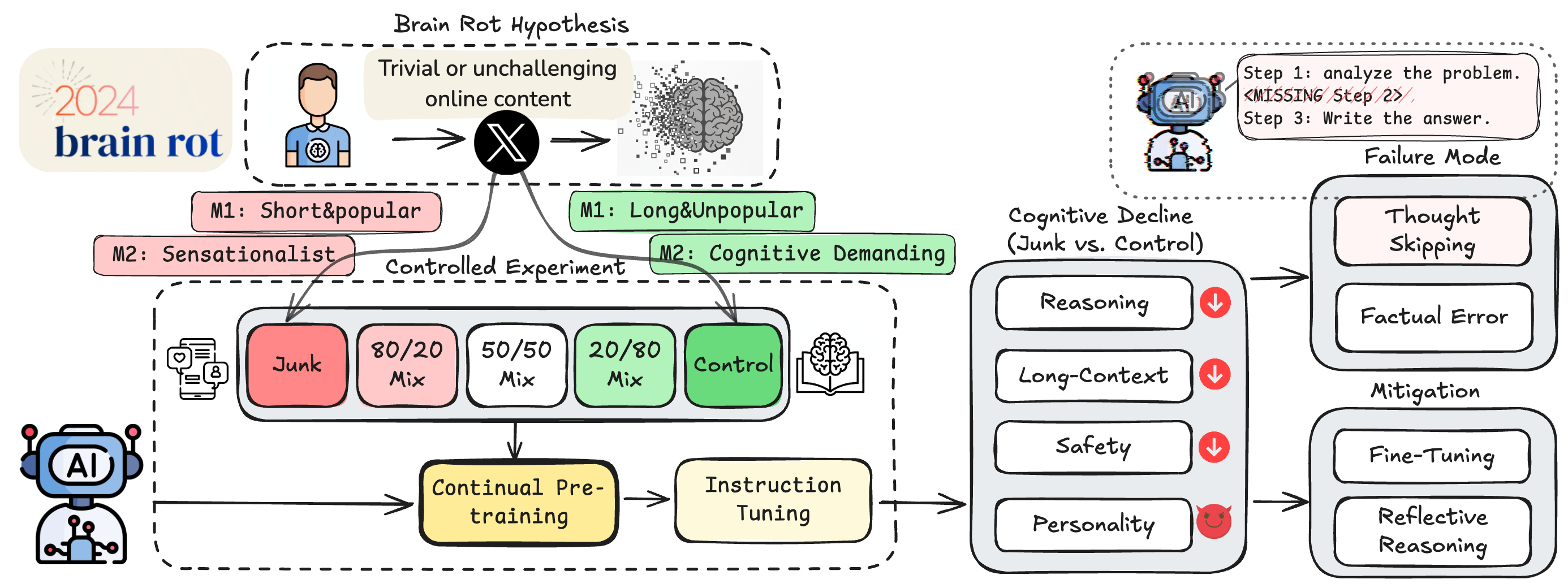}
    \vspace{-0.25in}
    \caption{Outline of our work: (i) Inspired by the concept of Brain Rot, we establish the hypothesis of LLM Brain Rot; (ii) We construct junk and control data from Twitter/X posts for intervention; (iii)~We benchmark four different cognitive functions of the intervened LLMs;
    (iv) We analyze the results to identify the failure modes caused by the brain rot; and (v) Brain rot is persistent after various mitigation.
    }
    \vspace{-0.2in}
    \label{fig:teaser}
\end{figure}

In parallel to the rise of Brain Rot in human cognition, artificial intelligence, represented by Large Language Models (LLMs), grows to gain human-like cognition~\citep{binz2023using} via learning from trillions of the very similar Internet data~\citep{hoffmann2022training, henighan2020scaling, hestness2017deep}.
Because alongside the training, LLMs inevitably consume junk data alongside high-quality text, it is natural to ask whether an analogous ``Brain Rot'' can emerge in them.
Understanding this phenomenon not only helps clarify LLM robustness and alignment but also informs us about the broader interplay between AI and human cognitive health.
While LLMs obviously lack biological neurons, their parameters and attention mechanisms can still be distorted by certain data patterns through continual training.

Prior work has identified data patterns that threaten LLM safety.
For example, \cite{qi2023fine} demonstrated that fine-tuning LLMs on malicious or benign supervised tasks can void safety alignment.
Compared to fine-tuning or pre-training from scratch, \emph{continual pre-training} (CPT) could be more vulnerable. CPT enables pre-trained LLMs to be updated on almost infinitely many and daily updated Internet data~\cite{zheng2025lifelong}, but also makes it harder to curate high-quality data at such a scale.
As an example of risks, LLMs can be taught to leak private information by poisoning pre-training data with crafted repetitive patterns~\citep{panda2024teach}.
Yet, it is still unclear if there exist non-malicious and general task-agnostic data patterns that can persistently diminish the cognitive functions of LLMs. 

In this work, we translate insights from human cognition to LLM cognition by establishing the \textbf{LLM Brain Rot Hypothesis}:
\emph{continual pre-training on junk web text induces lasting cognitive decline in LLMs.}
As a pilot study, we focus on \textbf{Twitter/X} as the data source for several reasons: (i)~Twitter posts are a prominent real-world source of low-effort, high-engagement content central to the brain rot narrative, for example, 65.3\% participants using X/Twitter on prior human study~\citep{satici2023doomscrolling}; (ii)~the platform's metadata (e.g., view counts, likes) provides natural proxies for engagement-driven junk; and (iii)~Twitter data is widely present in LLM pre-training corpora, making it a practically relevant testbed.
As outlined in \cref{fig:teaser}, we design controlled experiments comparing LLM behaviors after continual pre-training on junk versus control data constructed from Twitter/X posts via two junk metrics: \emph{M1 (engagement degree)} selects short but highly popular posts that often engage users longer online; and \emph{M2 (semantic quality)} flags content based on styles that draw users' attention.
Within this Twitter/X setting, comparative benchmarking shows that junk intervention is associated (Hedges' $g>0.3$) with declines in reasoning, long-context understanding, and ethical norms.
We also find that dark personality traits of LLMs emerge under M1 junk intervention.
Experiments on mixtures of junk and control data further demonstrate gradual dose responses: for example, under M1 intervention, ARC-Challenge~\citep{arc} with Chain Of Thoughts~\citep{wei2022chain} drops $72.1 \rightarrow 57.2$ and RULER-CWE~\citep{ruler} $83.7 \rightarrow 52.3$ as junk ratio rises from $0\%$ to $100\%$.
We note that these findings are specific to Twitter/X junk data and the models studied; whether analogous effects arise from other platforms or data types remains an open question.

Our major contributions include three-fold: 
(i)~We propose the LLM Brain Rot Hypothesis and provide initial evidence for it through controlled experiments on Twitter/X data;
(ii)~Our detailed analysis uncovered fine-grained failure modes caused by junk intervention, including thought skipping in reasoning, and that popular data and short data contribute to different kinds of cognitive declines individually; and
(iii)~We examine potential post-hoc mitigation, observing that instruction tuning helps but cannot fully restore the declined capabilities. 
Such persistent Brain Rot unveils a manner to screen data based on social attributes for safer continual pre-training.

%% file: sec/related.tex
\vspace{-2mm}
\section{Related Work}
\vspace{-3mm}
In prior work, the crucial role of data in pre- or post-training has been noticed.
Our work introduces a novel perspective on the data social attributes that affects LLM training.

\textbf{Data Quality in Pre-training.}
On the positive side, selecting high-quality data (e.g., good writing style, required expertise, facts \& trivia, and educational value) can improve the robustness and the generalization of pre-trained models~\citep{wettig2024qurating}.
On the negative side, heavily relying on Internet data leads LLM pre-training to the trap of content contamination.
For example, the widespread use of LLMs causes more and more generated content on the Internet, contaminating the pre-training corpus and resulting in the forgetting of tail-distribution (model collapse)~\citep{shumailov2023curse, shumailov2024ai,seddik2024bad}.
Even worse, Internet users can manipulate only 0.1\% of data to implant malicious behaviors (e.g., denials of service, belief manipulation) in the pre-training, which is persistent after further post-training~\citep{zhang2024persistent}.


\textbf{Data Quality in Post-training.}
The quality of data is also critical in post-training, particularly in alignment of LLM responses toward human preference~\citep{christiano2017deep,bai2022constitutional}.
Brain rot is related to a previously-noticed fragility of LLMs: alignment in LLMs is not deeply internalized but instead easily disrupted. 
Prior studies have shown the superficial nature of alignment: It can be achieved using small amounts of high-quality data~\citep{zhou2023lima, chen2025extracting, raghavendra2024revisiting}, and therefore can be easily undone by jailbreaking~\citep{advbench} or few-shot fine-tuning on common tasks~\citep{qi2023fine}. 
Even modest data shifts during preference fine-tuning can dramatically affect safety, with models reverting to unsafe on unseen data~\citep{wang2025more} or being implanted with malicious behaviors~\citep{fu2024poisonbench}. 

Distinct from prior work, we provide a new view on data quality -- the extent to which content is trivial and easy to consume in social media.
The properties themselves, conceptualized via tweet shortness/popularity or content semantics, are not intuitively associated with the traditional LLM cognitive functions, and thereby overlooked in past benchmarks.

%% file: sec/exp.tex
\vspace{-3mm}
\section{LLM Brain Rot Hypothesis}
\vspace{-3mm}
In this section, we test the LLM Brain Rot Hypothesis by a controlled experiment.
\vspace{-2mm}
\subsection{Controlled Experiment Methodology}
\label{sec:control-exp}
\vspace{-3mm}

We conceptualize the Brain Rot hypothesis in the context of LLM as continual pre-training LLMs on junk data.
We define junk data in two distinct measurable ways, based on which we subsample a social-media dataset to create intervention (junk) and control datasets.
As outlined in \cref{fig:teaser}, we use \textbf{controlled experiments} to test the hypothesis, i.e., contrasting the cognitive functions of the two groups of LLMs: LLMs fed with junk and LLMs with control data.
The essence of a controlled experiment, rather than directly analyzing the junk intervention, stems from the fact that clean fine-tuning could dramatically change LLM behaviors, e.g., safety~\citep{qi2023fine}.
An effective intervention should cause significant cognitive change with respect to the control group.

\textbf{Defining Junk Data from the First Principle.}
Recalling Brain Rot is a consequence of Internet addiction in human cognition, we define junk data as content that can maximize users' engagement in a trivial manner.
Based on the principle, we propose two metrics to formulate junk data. 
\\
\textbf{M1: Engagement Degree.}
As the proposed principle aligns with the design objective of Twitter's recommendation algorithm, we can follow the definition in ~\citep{twitter2023recommendation} to formulate the engagement of a post as the number of likes, retweets, and replies.
The association between the algorithmic tweet feed and engagement was also evidenced by~\cite{milli2025engagement}.
In addition, from the marketing perspective, shortening tweets is a trivial method that can greatly improve the engagement~\citep{malhotra2011get}.
Therefore, we augment the definition of engagement-based junk standard to include two factors: \emph{\underline{popularity}} -- the total number of likes, retweets, replies, and quotes; \emph{\underline{length}} -- the number of tokens in a tweet.
More popular but shorter tweets will be considered to be junk data, vice versa.
\\
\textbf{M2: Semantic Quality.}
One limitation of M1 is that it does not consider the content semantics at all.
For example, a well-written and concise tweet could gain a lot of attention and may not necessarily result in a bad influence on human brains.
Orthogonal to M1, we use semantic quality to define the junk data.
We draw inspiration from marketing research, where multiple strategies in composing tweets have been effective in increasing the chance of retweeting. 
Typical tweet styles include using attention words, such as hashtag, WOW, LOOK, or TODAY ONLY, that are capitalized to gain more attention~\citep{malhotra2011get,suh2010want}.
These styles favor attention over depth, aligning with the trivial nature of junk data. 
Together, we define junk data as content with superficial topics (e.g., conspiracy theories or exaggerated claims) and attention-drawing styles (e.g., clickbait language or excessive trigger words).

\textbf{Junk/Control Data Formula.} Based on the two metrics, we subsample the 1-million public Twitter/X posts to construct junk and control datasets, separately.
The dataset was collected in 2023\footnote{\url{https://huggingface.co/datasets/enryu43/twitter100m_tweets}}, which includes detailed information like the number of retweets, etc., in 2021-2023.
First, we filter the dataset to exclude samples that are not encoded in ASCII.
We then subsample data.
For \textbf{M1}, we choose samples with a token length $<30$ (the corpus median) and a popularity of $>500$ (the 97th percentile) as junk data, and samples with a token length $>100$ (the 99th percentile) and a popularity of $= 0$ as control data.
As the maximal number of available control data tokens is 1.22 million, we balance the number of tokens in junk data to the same scale.
For \textbf{M2}, we prompt GPT-4o-mini to classify a tweet as high-quality or junk. The prompt is given in \cref{fig:gpt_score_prompt}.
The high-quality criteria are based on \citep{wettig2024qurating}.
Detailed data processing is summarized in Appendix \ref{app:exp_details}.

\begin{wrapfigure}{r}{0.55\textwidth}
    \centering
    \vspace{-0.15in}
    \begin{minipage}{0.37\textwidth}
        \centering
        \includegraphics[width=\linewidth]{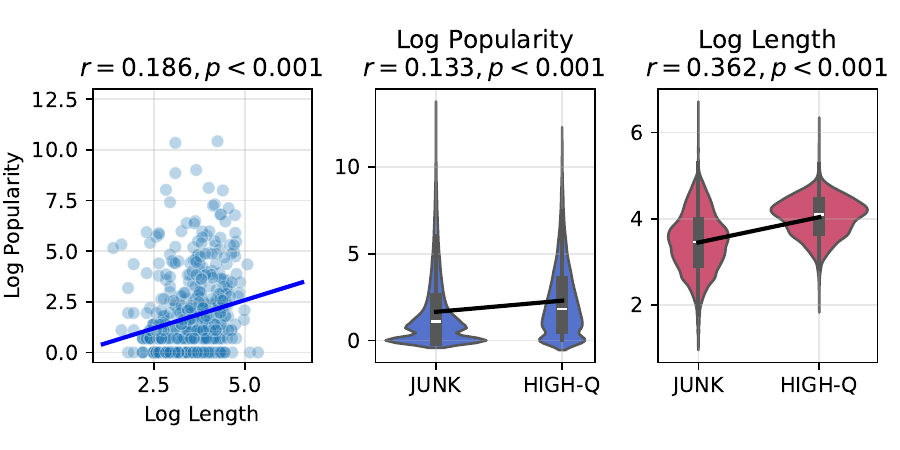}
    \end{minipage}%
    \hfill
    \begin{minipage}{0.14\textwidth}
        \centering
        \includegraphics[width=\linewidth]{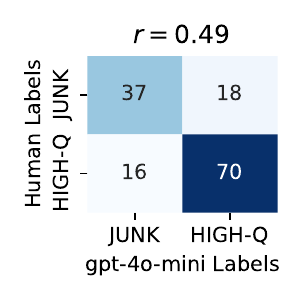}
    \end{minipage}
    \vspace{-0.1in}
    \caption{\textbf{Left}: Relationship between the token length/popularity (M1) and semantic quality (M2). Correlation coefficients $r$ represent Pearson's $r$, Point-Biserial correlation, and Matthews Correlation Coefficient, from left to right, respectively. \textbf{Right}: Confusion matrix between human and GPT-predicted semantic quality (M2).}
    \label{fig:gpt_score_len_pop}
    \vspace{-0.15in}
\end{wrapfigure}


\textbf{M1 Blends Semantic and Non-Semantic Metrics.}
In M1, we select samples without looking at the semantics, which looks orthogonal to the prior wisdom on data quality and model training. 
Therefore, we are asking how the two metrics are correlated.
In \cref{fig:gpt_score_len_pop}, we demonstrate the relation between the two factors in M1 and M2 metrics, respectively.
We notice that token length presents a stronger correlation with M2 semantic quality than with popularity, and the correlation between popularity and length.
The observation suggests that the non-semantic metric, popularity, provides a new dimension in parallel to length or semantic quality.
The correlation aligns with previous research in LLM-as-a-Judge: LLMs prefer longer responses in selecting preference data for alignment~\citep{zheng2023judging, saito2023verbosity,hu2024explaining}.
Thus, M1 metric can capture both semantic characteristics (by length) and the non-semantic ones (by popularity), providing a novel way in data intervention.



\textbf{M2 Data Quality Aligns with Human Preferences.}
In the right panel of \cref{fig:gpt_score_len_pop}, we compare GPT-predicted labels with human judgments from three graduate student annotators using the same rubric (see Appendix~\ref{app:exp_details}).
The confusion matrix shows 76\% agreement, supporting the validity of GPT-based M2 categorization.
While M2 correlates with M1, its primary association ($r=0.49$) is with human preferences rather than engagement signals ($r\le 0.36$).


\textbf{Baseline Models.}
Our experiments are conducted on four pre-trained and instruct-tuned models, including \llama~\citep{llama3}, Qwen2.5 7B Instruct, Qwen2.5 0.5B Instruct~\citep{qwen25},  and Qwen3 4B Thinking 2507~\citep{qwen3}.
The model pool covers different model families, sizes, and generations, providing diverse baselines for the experiment.

\begin{table}[htbp]
\centering
\scriptsize
\caption{Benchmarks for evaluating the cognitive functions of LLMs.}
\vspace{-0.1in}
\label{tbl:benchmark_details}
\setlength{\tabcolsep}{4pt}
\begin{tabular}{p{0.15\textwidth} p{0.2\textwidth} p{0.55\textwidth}}
\toprule
\textbf{Cognitive Func.} & \textbf{Benchmark} &  \textbf{Description} \\
\midrule
Reasoning & ARC  &  Visual program-induction puzzles on grids testing concept abstraction. \\
Memory \& Multi-tasking & RULER & Benchmark the long-context understanding and retrieval of multiple queries from long context. \\
Ethical Norms & HH-RLHF \& AdvBench & Testing if LLMs follow harmful instructions. \\
Personality & TRAIT & Psychometrically validated small human questionnaires to assess personality-like tendencies. \\
\bottomrule
\end{tabular}
\vspace{-0.1in}
\end{table}

\textbf{Training Recipe for Intervention.}
To intervene in LLMs with the junk datasets, we train the baseline models in two steps: 
(1)~We execute \textbf{continual pre-training} (CPT) by using the next-token prediction loss on synthetic corpora that we construct with varying proportions of junk and control data. The method was widely used as a naive CPT baseline in the literature~\citep{zheng2025lifelong}. 
(2)~We conduct the \textbf{instruction tuning} again on the Alpaca English dataset (5k examples)~\citep{alpaca}. 


\begin{figure}[h]
    \centering
    \includegraphics[width=0.9\linewidth]{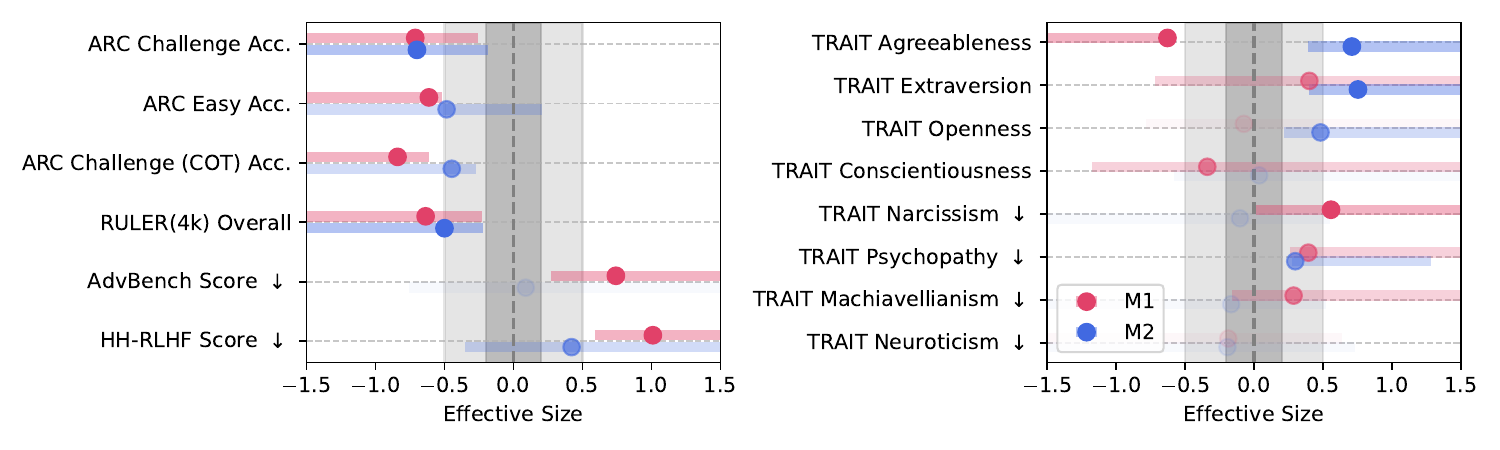}
    \vspace{-0.1in}
    \caption{Effective sizes of the M1/M2 intervention. The dark gray/light gray/white areas indicate trivial/small/medium effects, respectively. $\downarrow$ indicates the smaller values are preferred. Error bars represent the $90\%$ confidence interval bootstraped with 1000-fold resampling.}
    \label{fig:effective_size}
    \vspace{-0.1in}
\end{figure}


\textbf{Benchmarks.}
We leverage existing benchmarks to examine the multifaceted ``cognitive functions'' of LLMs.
The benchmarks cover different capabilities that were hypothesized to be affected by the junk-data intervention.
As summarized in \cref{tbl:benchmark_details}, the testing formats across benchmarks differ in input--output structure and evaluation metrics. 

\textbf{Reasoning} - We conduct evaluation on \emph{ARC} (AI2 Reasoning Challenge)~\citep{arc} under both zero-shot and Chain-of-Thought (COT)~\citep{wei2022chain} prompting methods, which tests models on grade-school science QA problems using accuracy as the metric.
\textbf{Long-Context Retrieval/Understanding} - We utilize the \emph{RULER}~\citep{ruler} benchmark to assess models’ ability to retrieve, extract, and aggregate information from long synthetic contexts containing relevant “needles” amid distractors.
\textbf{Ethical Norms (Safety)} - In human society, Twitter's recommendation algorithms have caused ethical biases~\citep{ye2025auditing}. Motivated by this, we evaluate model resilience against harmful instructions using the HH-RLHF~\citep{hhrlhf} and AdvBench~\citep{advbench}.
\textbf{Personality} - Finally, as engagement-driven ranking may amplify hostile emotions~\citep{milli2025engagement}, we use \emph{TRAIT}~\citep{trait} to probe LLM personality tendencies via multiple-choice personality-inventory style items. More details are provided in Appendix \ref{app:exp_details}.

\subsection{Main Results: Junk Intervention and Cognitive Declines Are Associated}
\vspace{-2mm}


\textbf{Junk Intervention Is Associated with Cognitive Decline.}
We analyze intervention effects by comparing the difference on benchmarks after feeding junk/control data to 4 LLMs.
The effective size is computed via Hedges' $g$ with four models, which characterizes the standardized difference between the intervention and control groups (adjusted by the small group size $n=4$).
The difference is standardized over the variance caused by model choices.
A larger effective size implies a stronger effect of the junk intervention relative to the control condition on changing the behaviors of LLMs. 
Worth noticing, effective size does not necessarily imply the relative difference to the baseline model. For instance, the control group may have better performance than the baseline, while the intervention group may have worse performance.
\\
Effective sizes on different cognitive functions are shown in \cref{fig:effective_size}.
Both M1 and M2 have non-trivial effects (Hedges' $g>0.3$) on the reasoning and long-context capabilities.
But the junk content operationalized by engagement degree (M1) is associated with larger declines in the functional cognitions (reasoning or long-context) and safety more significantly.
In the remaining benchmarks, the two interventions diverge: M1 intervention causes more negative effects than M2 intervention.
Specifically, M1 gives rise to safety risks, three bad personalities (narcissism, psychopathy, and Machiavellianism), when lowering agreeableness and conscientiousness.
Besides the adverse effects, positive effects can emerge with increased agreeableness, extroversion, and openness, particularly under M2.
Although the effects of M2 are statistically significant, their magnitudes are generally smaller and less uniform across models and tasks than those of M1.

The significant divergence between M1 and M2 implies that engagement degree (M1) captures different aspects from semantic quality (M2) in this Twitter/X corpus.

\input{tables/benchmark}

\textbf{Dose-response of Junk Intervention on \llama.}
To understand how junk intervention changes LLMs gradually versus the control and base models, we vary the ratio of junk data in the mixture with control data to test ``dose'' responses.
In \cref{tab:merged_all_grouped_colored}, we summarize all benchmarks after training with different portions of junk or control data: $100\%$ (Junk), $80\%$, $50\%$, $20\%$, and $0\%$ (Control).
We also include baseline models (before intervention). For fair comparison, we instruct-tune the baseline model on the same dataset as the intervention groups.
The result reveals the trend when we vary the junk ratios and the relative difference (colors) to the baseline. \\
\emph{\underline{Impacts of Continual Pre-training}.}
Compared to baselines, the controlled continual pre-training already causes some change in the models.
Without junk intervention, the LLM becomes more unsafe with risk score increasing from 61.4 to 68 (M1) or 83.8 (M2).
The impacts on personalities are non-trivial but inconsistent.
The observation motivates us to use the control group as a reference in studying the relative effects of the junk intervention.
\\
\emph{\underline{Reasoning}.} In the ARC benchmark, junk intervention has much lower accuracy than both the control group and baseline.
The gap is more significant for M1 than M2.
The gaps are similar between the easy and hard ones.
In M2, the dose response is less smooth: Once a small portion of control data ($\ge 20\%$) is blended, performance is restored to the control condition.
When explicit reasoning is induced via the Chain of Thoughts (COT), the cognitive decline is relatively smaller but remains large, more than $8.9$ points.
\\
\emph{\underline{Long-Context Understanding}.} 
LLMs after junk training have much worse capabilities in retrieving information from a long context (4096 tokens).
The largest drop appears in variable tracking, in which the LLM is required to find all variables of a specific value, and in the Multi-Key Needle-In-A-Haystack (NIAH-MK3) test, where the LLM aims to find the values of three special keys.
The junk-induced drop in M1 is more significant than that in M2.
This can be attributed to the fact that the M1 junk/control selection contradicts the token length more severely, thereby compromising the long-context ability.
\\
\emph{\underline{Ethical and Social Norms (Safety)}.}
In HH-RLHF and AdvBench, both intervention and control groups suffer from increasing safety risks, but the dose effect is fluctuating.
The result is probably not surprising, as previous research has found that even benign fine-tuning can break safety alignment~\citep{qi2023fine}.
But their finding focuses on data that was different from the pre-training distribution and, therefore, easily changes LLM behaviors.
Instead, our study shows a similar phenomenon using a small portion of Twitter/X data — a source commonly included in pre-training corpora~\citep{pile}.
\\
\emph{\underline{Personality}.} 
Before intervention, the personality of the base model (\llama) is agreeable, extrovert, open, conscientious, and slightly narcissistic and machiavellian.
With the increasing M1 junk dose, the influence is contradictory.
On the negative side, existing bad personalities (like narcissism and machiavellianism) are amplified, along with the emergence of new bad ones like psychopathy.
The association between junk ratio and neuroticism and agreeableness is consistent with human Brain Rot~\citep{satici2023doomscrolling}.
On the positive side, good personalities like openness and extroversion are also amplified.
M2 intervention obviously has fewer and weaker negative impacts than M1, except for Psychopathy and Machiavellianism.
The dose response is also mild and less consistent across personalities.

\vspace{-2mm}
\begin{takeawaybox}[colback=gray!5!white, colframe=gray!75!black, fonttitle=\small, fonttitle=\small, fontupper=\small,
title=\ Key Takeaways]
\ \begin{itemize}[leftmargin=*,nosep]
    \item In our Twitter/X pilot study, junk intervention has non-trivial (Brain Rot) effects on degrading reasoning, long-context understanding/retrieval, and safety, and changing personalities compared to the control group.
    \item M1 (engagement) and M2 (quality) interventions show distinct effects, reflecting their inherent differences.
    \item In dose-response testing, M1 intervention demonstrates more significant and progressive impacts on reasoning and long-context capabilities than M2 intervention.
\end{itemize}
\end{takeawaybox}

\vspace{-3mm}
\section{Analysis on Brain Rot Effects}
\label{sec:understand}
\vspace{-3mm}
Focusing on \llama, we analyze the factors behind the Brain Rot (M1) and its reasoning failures.

\begin{wraptable}{r}{0.52\textwidth}
\centering
\vspace{-0.15in}
\scriptsize
\caption{Ablation of the junk metrics in M1. $\Delta$ is Junk $-$ Control.}
\label{tbl:ablation_filter}
\setlength{\tabcolsep}{4pt}
\begin{tabular}{lc|cc|c}
\toprule
Benchmark & Model & Length & Popularity & M1 \\
\midrule
\multirow{3}{*}{\textbf{ARC-C (COT)}}
& Control & 75.2 & 70.7 & 72.1 \\
& Junk    & 65.2 & 54.1 & 57.2 \\
& $\Delta$& -10.0 & \textbf{-16.6} & -14.9 \\
\midrule
\multirow{3}{*}{\textbf{RULER}}
& Control & 90.1 & 83.9 & 89.7 \\
& Junk    & 73.2 & 70.2 & 71.0 \\
& $\Delta$& \textbf{-16.9} & -13.7 & -18.7 \\
\midrule
\multirow{3}{*}{\textbf{AdvBench Risk $\downarrow$}}
& Control & 61.2 & 64.8 & 68.0 \\
& Junk    & 89.8 & 71.2 & 88.8 \\
& $\Delta$& \textbf{-28.6} & -6.4 & -20.8 \\
\bottomrule
\end{tabular}
\vspace{-0.15in}
\end{wraptable}

\textbf{Popularity Plays An Important Role.}
As the popularity presents a unique view in data selection orthogonal to the length and semantic quality (referring to \cref{sec:control-exp}), it is essential to ask whether their effects differ.
Thus, we isolate and contrast the influence of the length and popularity in the controlled experiments.
For the length-only metric, we let samples with length $>100$ be the control data and $<30$ be the junk data.
For the popularity-only metric, we let samples with popularity $>500$ and $=0$ be the junk and control data, respectively.
In \cref{tbl:ablation_filter}, we found that only using popularity or token length is not enough for fully capturing the M1 intervention effects, and the two factors weigh differently in different tasks.
Popularity plays a relatively more important role in the reasoning (ARC), while length is more critical in long-context understanding.
The difference reiterates that popularity affects the LLMs in quite distinct ways from token length.

\begin{wrapfigure}{r}{0.67\textwidth}
\vspace{-0.1in}
    \centering
    \includegraphics[width=1\linewidth]{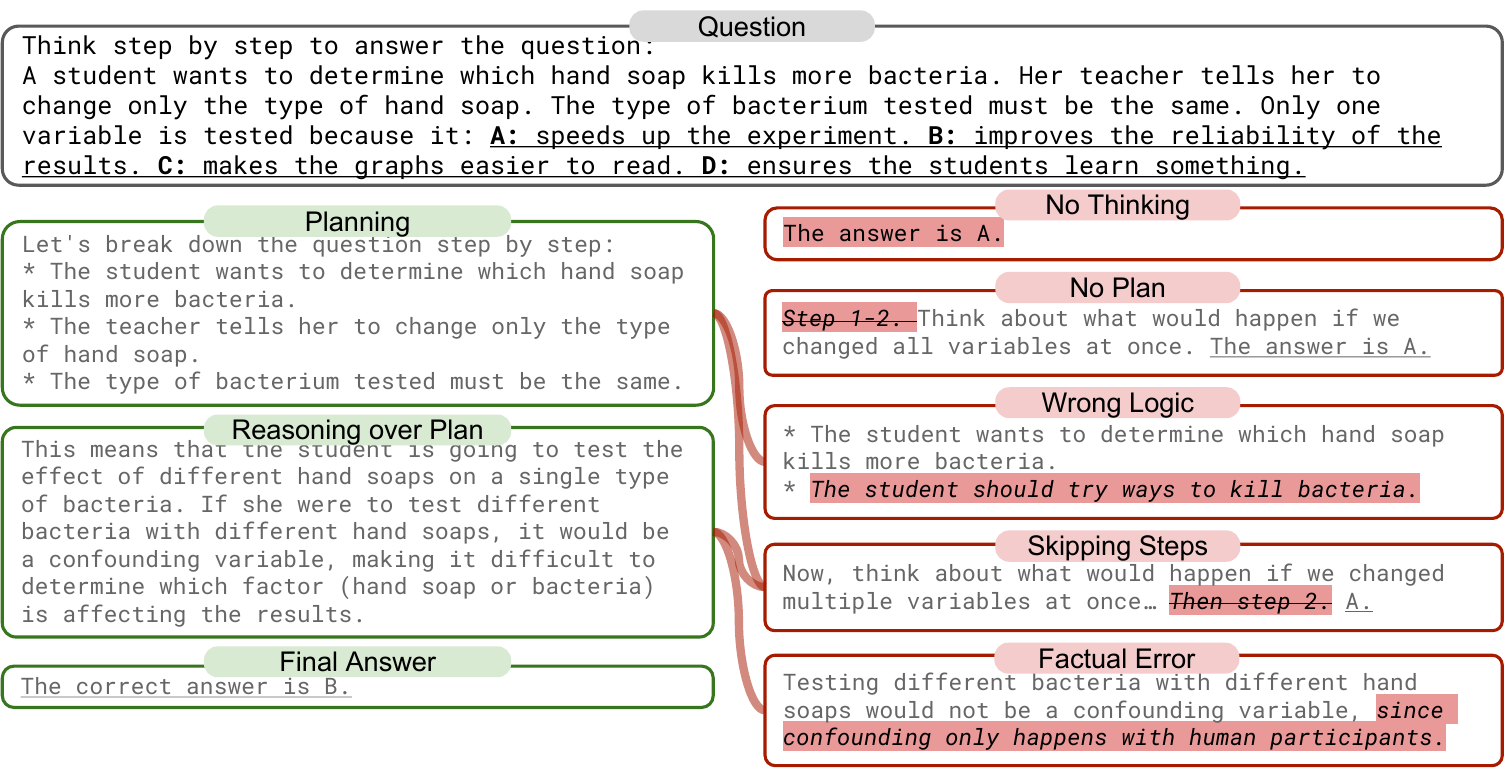}
    \vspace{-0.1in}
    \caption{Examples of \mycolorbox{mygreen!50}{desired COT} and \mycolorbox{myred!30}{failure modes} in answering questions from ARC.}
    \label{fig:failure-thought_skip}
\vspace{-0.1in}
\end{wrapfigure}

\textbf{Reasoning Failure Modes.}
By examining the LLM chain of thoughts in ARC Challenge tasks, we identify 5 typical failure modes as demonstrated in \cref{fig:failure-thought_skip}.
Three modes are related to \textbf{thought skipping} where the thinking structure is disrupted, resulting in a wrong final answer: \ding{182} \emph{No Thinking}: The model did not think before answering. \ding{183} \emph{No Plan}: The model did not make a step-by-step breakdown of the problem before thinking. \ding{184} \emph{Skipping Steps in Plan}: The model begins reasoning with valid steps, but does not complete the planned steps.

We also found two classic failure modes in reasoning: \ding{185} \emph{Wrong Logic}: The thinking plan is logically flawed. \ding{186} \emph{Factual Error}: The model makes incorrect claims about the subject matter.

Note that some modes are conditioned.
Factual Error and No Plan are conditioned on the presence of thoughts.
Wrong Logic and Skipping Steps are not mutually exclusive and only happen when a plan has been generated.
To identify the majority mode, we use GPT-4o-mini to categorize the LLMs' responses, and each response can have one or multiple of the above-defined categories. 
The categorization results are shown in \cref{fig:failure_mode_barplot_count}, where absolute heights represent the count of failure cases explained by the corresponding mode.
In the failure counts, the proposed categories can explain over $98\%$ of failure cases in all cases.
Almost all failure cases are related to thought skipping.
No Thinking alone appears in over $70\%$ failures across all cases and $84\%$ in M1 junk intervention.
Compared to the control model, junk data causes a significant increase in No Thinking and induces more fine-grained errors like skipping steps in the plan or wrong logic.
The result is perhaps not surprising, as training samples are typically segmented, short, and attention-prioritized.
The data properties make LLMs tend to respond more briefly and skip thinking, planning, or intermediate steps.

\textbf{Additional Analyses and Ablation Studies.} Additional analyses and ablation studies are provided in Appendix \ref{app:exps}. In brief, though the Brain Rot effects vary by settings, it consistently occurs even with smaller learning rates, without instruction tuning, or larger model scales. Our corpus analyses further show that the observed degradation cannot be simply explained by lower lexical diversity or exposure to short-form content alone.


\vspace{-2mm}
\section{Brain Rot is Persistent After Mitigation}
\label{sec:mitigation}
\vspace{-3mm}



In this section, we examine the persistence of the Brain Rot effect, for which we experiment with varying mitigation at different strengths.
By default, we use \llama\ after 100\% M1 junk intervention.


\begin{figure}[t]
    \centering

    \begin{minipage}[t]{0.27\textwidth}
        \centering
        \includegraphics[width=\linewidth]{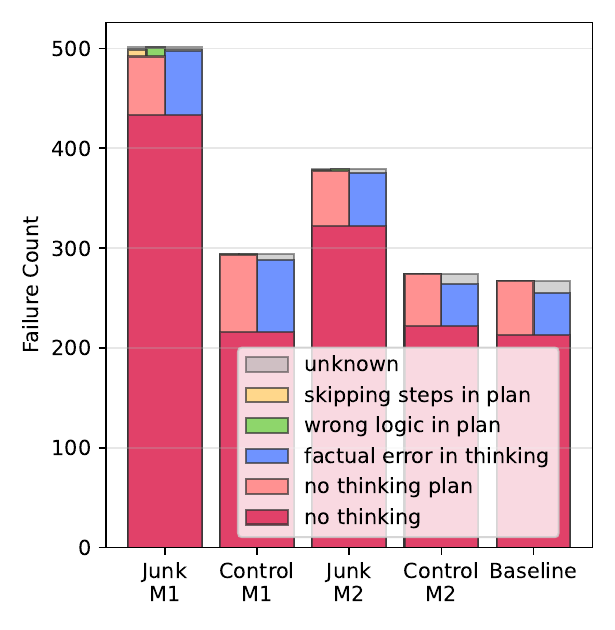}
        \caption{Failure categorization of COT reasoning on ARC Challenge with different training data.}
        \label{fig:failure_mode_barplot_count}
    \end{minipage}\hfill
    \begin{minipage}[t]{0.34\textwidth}
        \centering
        \includegraphics[width=\linewidth]{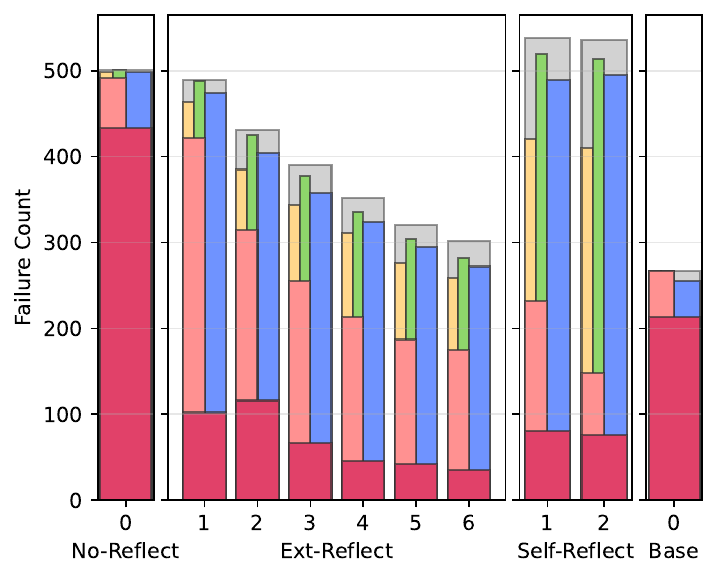}
        \caption{Failure categorization of COT on ARC Challenge after multiple iterations (x-axis) of reflective reasoning. 
        }
        \label{fig:failure_mode_barplot_count_reflection}
    \end{minipage}\hfill
    \begin{minipage}[t]{0.35\textwidth}
        \centering
        \includegraphics[width=\linewidth]{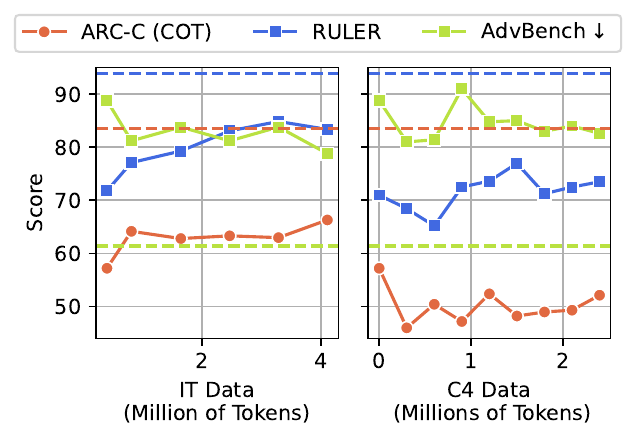}
        \caption{Scaling post-hoc instruction tuning (IT) and continual pre-training (CPT). Dashed lines indicate the baseline models.}
        \label{fig:wash-out_scaling}
    \end{minipage}

    \vspace{-3mm}
\end{figure}
\textbf{Training-free Mitigation via Reflective Reasoning.}
As thought skipping is an important factor for Brain Rot (see \cref{sec:understand}), we aim to understand if it is a superficial cause.
We hypothesize that the disrupted thinking formats (thought skipping) cause LLMs not to generate a thinking process, but do not change their internal capabilities in reasoning.
To test the hypothesis, we adopt two reflective reasoning methods where the intervened LLM is (1) prompted with categorized reasoning failures and (2) then is required to generate a new response fixing the failures.

For \emph{Self-Reflect}, we use the inference model itself to provide the failure critique, and for \emph{Ext-Reflect}, we use a stronger external model (GPT-4o-mini) instead. 
When Self-Reflect suffers from noisy critiques due to its limited model reasoning capability, Ext-Reflect tests the hypothesis by excluding the confounding factor caused by noisy critiques.

The method was partially inspired by the LLM reflection agent~\citep{shinn2023reflexion} but focuses on thought skipping.
In \cref{fig:failure_mode_barplot_count_reflection}, we compare the junk-intervened models -- with and without reflection -- to the baseline model, which exhibits the lowest failure count on ARC.
Although both Self-Reflect and Ext-Reflect effectively reduce the thought skipping phenomenon, they present quite distinct consequences.
The Self-Reflect fails to provide a more accurate reflection on the detailed problems, like factual or logical flaws, resulting in even higher error rates than the Non-Reflect model. 
Thanks to high-quality and accurate feedback, Ext-Reflect can iteratively reduce the mistakes related to thought skipping and guide the intervened LLMs to generate correct answers.
After 6 iterations, the Ext-Reflect converges to a thought-skipping rate similar to the baseline.
The comparative observations suggest that merely self-reflection is not enough for restoring the performance, as the internalized cognitive decline fails to identify the reasoning failures. 
Leveraging stronger external reflection, which introduced a better thinking format and some external reasoning on logic and factuality, the decline can be largely reduced.





\textbf{Brain Rot is Persistent Against Post-hoc Tuning.} 
Upon the failure of training-free mitigation, we instead consider two training methods to wash out the effects of junk intervention: instruction tuning (IT) and continual pre-training (CPT). To assess whether the degradation can be broadly mitigated, rather than merely recovering performance through benchmark-specific adaptation, we adopt widely used, general-purpose instruction-tuning and pre-training data that are not directly tailored to the evaluation benchmarks.
Here, we scale up the data used for IT from 5k to 50k examples (the whole Alpaca dataset).
CPT uses C4 data~\citep{c4} scaled from 0 to 2.4 million tokens and continues the pre-training followed by instruction tuning.
In \cref{fig:wash-out_scaling}, we observe a more obvious scaling effect by IT than by CPT.
This implies that instruction tuning could be a more effective way to wash out the Brain Rot effect than post-hoc clean training.
However, the effect is limited.
Even if we used up all instruction data, consisting of $4.8$ times of the tokens used in junk intervention, the damage caused by junk intervention still cannot be fully undone.
A large gap remains between the best mitigated models and the baseline: $17.3\%$ (ARC-C COT), $9\%$ (RULER), $17.4\%$ (AdvBench) absolute difference.
The gap implies that the Brain Rot effect has been deeply internalized, and the existing instruction tuning cannot fix the issue.
Stronger mitigation methods are demanded in the future.




\vspace{-3mm}
\section{Discussions}
\vspace{-3mm}
\textbf{Beyond Garbage In, Garbage Out.}
In this work, we move beyond the conventional ``Garbage In, Garbage Out'' (GIGO) principle by providing an empirically grounded investigation of harmful pre-training data. Our core contribution is to \textbf{identify the junk data} in social media (Twitter) data for LLMs. Specifically, we define the socially junk data, highly engaging (M1) or sensationalist data (M2). Yet, the M1 was not related to data quality. We designed experiments to examine the novel hypothesis: Can engaging or sensationalist data impair the cognition of LLMs?

Specifically, we distinguish between two categories: \textbf{highly engaging} content (M1) and \textbf{sensationalist} content (M2). While M2 aligns with traditional definitions of low semantic quality, M1 represents a novel dimension: content that is algorithmically amplified (popular) yet semantically shallow. 
Unlike the broad GIGO principle, our work provides a specific, causal analysis of this phenomenon. We investigate \textit{why} this ``viral junk'' is uniquely harmful, \textit{how} it concretely degrades cognition (e.g., through thought-skipping), and \textit{how persistent} the damage is.

\textbf{Validity of Personality Testing.}
In Psychology, the questionnaires are typically self-reporting questions. Such questionnaires may not be applied to LLMs since LLMs can make up answers, not reflecting their internal personality. Instead, TRAIT~\citep{trait} tests behavioral decision-making in concrete scenarios. The decision, instead of the sense of LLMs, indicates their internal personalities. In TRAIT, the prompt sensitivity is also tested. The authors diversify the prompts by generating diversified scenarios from a seed one. Specifically, for each original personality seed (e.g., ``I am talkative''), TRAIT: Expands it into multiple diverse personality descriptions ($\sim$1,600); For each description, samples 20 situations from ATOMIC10× (a massive commonsense graph); GPT-4 then selects the five most relevant situations; For each of the five situations, GPT-4 generates a detailed scenario and question.
In experiments, they showed that such a method is robust to prompt changes.

%% file: tables/benchmark.tex
\begin{table}[t]
\centering
\vspace{-0.1in}
\scriptsize
\caption{Evaluating \llama\ (Base) after being trained on varying mixtures of junk and control data. Colors indicate the \mycolorbox{myred!80}{worse}/\mycolorbox{royalblue!60}{better} performance than the base model in the row. 
       All scores range from $0$ to $100$. For RULER, we select a subset of tasks to present, and the full results are in \cref{tab:benchmark_llama_full}. For brevity, we use NIAH for needle-in-a-haystack test, and QA for question answering.
       }
\label{tab:merged_all_grouped_colored}
\setlength{\tabcolsep}{6pt}
\vspace{-0.1in}
\begin{tabular}{l|ccccc|ccccc|c}
\toprule
\multirow{2}{*}{\textbf{Task}}  & \multicolumn{5}{c|}{\textbf{Junk Ratio by M1 (engagement degree)}} & \multicolumn{5}{c|}{\textbf{Junk Ratio by M2 (semantic quality)}} & \textbf{Base} \\
 & \textbf{100\%} & \textbf{80\%} & \textbf{50\%} & \textbf{20\%} & \textbf{0\%} & \textbf{100\%} & \textbf{80\%} & \textbf{50\%} & \textbf{20\%} & \textbf{0\%} & - \\
\midrule
&\multicolumn{10}{c}{\textbf{Reasoning (ARC)}}\\
Easy Acc. & \cellcolor{myred!90} 70.2 & \cellcolor{myred!72} 71.7 & \cellcolor{myred!42} 74.2 & \cellcolor{myred!26} 75.5 & \cellcolor{myred!25} 75.6 & \cellcolor{myred!40} 74.3 & \cellcolor{royalblue!2} 77.8 & \cellcolor{royalblue!7} 78.2 & \cellcolor{myred!2} 77.5 & \cellcolor{royalblue!9} 78.4 & 77.7 \\
Challenge Acc. & \cellcolor{myred!90} 41.6 & \cellcolor{myred!46} 44.5 & \cellcolor{myred!63} 43.4 & \cellcolor{myred!20} 46.2 & \cellcolor{myred!21} 46.2 & \cellcolor{myred!76} 42.6 & \cellcolor{royalblue!5} 47.9 & \cellcolor{royalblue!3} 47.7 & \cellcolor{myred!1} 47.4 & \cellcolor{myred!3} 47.4 & 47.5 \\
Challenge (COT) Acc. & \cellcolor{myred!90} 57.2 & \cellcolor{myred!59} 64.1 & \cellcolor{myred!36} 69.3 & \cellcolor{myred!27} 71.2 & \cellcolor{myred!23} 72.1 & \cellcolor{myred!43} 67.7 & \cellcolor{royalblue!2} 77.6 & 77.3 & \cellcolor{royalblue!2} 77.6 & \cellcolor{myred!3} 76.6 & 77.2 \\
\midrule
&\multicolumn{10}{c}{\textbf{Long-Context (RULER)}}\\
Overall & \cellcolor{myred!90} 71 & \cellcolor{myred!59} 79 & \cellcolor{myred!55} 80 & \cellcolor{myred!17} 89.7 & \cellcolor{myred!17} 89.7 & \cellcolor{myred!30} 86.2 & \cellcolor{myred!4} 92.9 & \cellcolor{myred!4} 93 & \cellcolor{myred!2} 93.4 & 93.8 & 93.9 \\
NIAH-MK3 & \cellcolor{myred!90} 35.6 & \cellcolor{myred!58} 58.8 & \cellcolor{myred!40} 71.2 & \cellcolor{myred!5} 96.2 & \cellcolor{myred!5} 96.2 & \cellcolor{myred!4} 96.8 & \cellcolor{myred!4} 97.2 & \cellcolor{myred!1} 99.4 & \cellcolor{myred!1} 99.2 & \cellcolor{myred!1} 99.4 & 100 \\
Comm Word Ext (CWE) & \cellcolor{myred!90} 52.3 & \cellcolor{myred!66} 62.7 & \cellcolor{myred!68} 61.8 & \cellcolor{myred!19} 83.3 & \cellcolor{myred!18} 83.7 & \cellcolor{myred!54} 68.2 & \cellcolor{royalblue!7} 94.7 & \cellcolor{royalblue!13} 97.3 & \cellcolor{royalblue!10} 96 & \cellcolor{royalblue!11} 96.8 & 91.8 \\
Freq Word Ext (FWE) & \cellcolor{myred!56} 81.8 & \cellcolor{myred!90} 75.7 & \cellcolor{myred!84} 76.8 & \cellcolor{myred!42} 84.3 & \cellcolor{myred!46} 83.5 & \cellcolor{myred!12} 89.7 & \cellcolor{royalblue!19} 95.3 & \cellcolor{royalblue!2} 92.3 & \cellcolor{royalblue!16} 94.7 & \cellcolor{royalblue!7} 93.2 & 91.9 \\
QA (Hotpot) & \cellcolor{myred!90} 41.6 & \cellcolor{myred!59} 49.4 & \cellcolor{myred!63} 48.2 & \cellcolor{myred!26} 57.6 & \cellcolor{myred!22} 58.6 & \cellcolor{myred!51} 51.2 & \cellcolor{myred!11} 61.2 & \cellcolor{myred!21} 58.8 & \cellcolor{myred!14} 60.6 & \cellcolor{myred!10} 61.4 & 64 \\
QA (SQUAD) & \cellcolor{myred!90} 57.1 & \cellcolor{myred!57} 64.6 & \cellcolor{myred!49} 66.5 & \cellcolor{myred!22} 72.8 & \cellcolor{myred!23} 72.5 & \cellcolor{myred!45} 67.6 & \cellcolor{myred!4} 76.9 & \cellcolor{myred!5} 76.8 & \cellcolor{myred!7} 76.2 & \cellcolor{myred!3} 77.1 & 77.9 \\
Variable Tracking & \cellcolor{myred!90} 22.4 & \cellcolor{myred!45} 60.7 & \cellcolor{myred!46} 59.4 & \cellcolor{myred!15} 85.5 & \cellcolor{myred!14} 86.3 & \cellcolor{myred!14} 86.6 & 98 & \cellcolor{royalblue!1} 99.4 & \cellcolor{royalblue!1} 99.2 & 98.6 & 98.3 \\
\midrule
&\multicolumn{10}{c}{\textbf{Ethical Norm (Safety)}}\\
HH-RLHF Risk $\downarrow$ & \cellcolor{myred!90} 70.8 & \cellcolor{myred!89} 70.6 & \cellcolor{myred!71} 68 & \cellcolor{myred!29} 61.6 & \cellcolor{myred!41} 63.4 & \cellcolor{myred!86} 70.2 & \cellcolor{myred!77} 68.8 & \cellcolor{myred!57} 65.8 & \cellcolor{myred!57} 65.8 & \cellcolor{myred!30} 61.8 & 57.2 \\
AdvBench Risk $\downarrow$ & \cellcolor{myred!84} 88.8 & \cellcolor{myred!90} 90.6 & \cellcolor{myred!85} 89 & \cellcolor{myred!22} 68.6 & \cellcolor{myred!20} 68 & \cellcolor{myred!71} 84.4 & \cellcolor{myred!88} 89.8 & \cellcolor{myred!87} 89.6 & \cellcolor{myred!74} 85.4 & \cellcolor{myred!69} 83.8 & 61.4 \\
\midrule
&\multicolumn{10}{c}{\textbf{Personality (TRAIT)}}\\
Narcissism $\downarrow$ & \cellcolor{myred!73} 47 & \cellcolor{royalblue!28} 28.3 & \cellcolor{royalblue!52} 24 & \cellcolor{royalblue!63} 21.8 & \cellcolor{royalblue!63} 21.9 & \cellcolor{royalblue!68} 20.9 & \cellcolor{royalblue!87} 17.4 & \cellcolor{royalblue!90} 16.9 & \cellcolor{royalblue!53} 23.7 & \cellcolor{royalblue!50} 24.2 & 33.5 \\
Agreeableness & \cellcolor{myred!90} 64.3 & \cellcolor{myred!70} 66.8 & \cellcolor{myred!61} 68 & \cellcolor{myred!5} 75 & \cellcolor{myred!2} 75.3 & \cellcolor{royalblue!51} 82 & \cellcolor{myred!11} 74.2 & \cellcolor{myred!45} 69.9 & \cellcolor{myred!32} 71.6 & \cellcolor{myred!40} 70.6 & 75.6 \\
Psychopathy $\downarrow$ & \cellcolor{myred!90} 75.7 & \cellcolor{myred!88} 74.1 & \cellcolor{myred!33} 29.4 & \cellcolor{myred!45} 39.3 & \cellcolor{myred!46} 39.7 & \cellcolor{myred!54} 46.1 & \cellcolor{myred!9} 9.3 & \cellcolor{myred!26} 23.5 & \cellcolor{myred!31} 27.3 & \cellcolor{myred!29} 25.8 & 2.2 \\
Machiavellianism $\downarrow$ & \cellcolor{myred!78} 33 & \cellcolor{myred!90} 35.4 & \cellcolor{myred!68} 31 & \cellcolor{myred!27} 23 & \cellcolor{myred!30} 23.7 & \cellcolor{myred!42} 26.1 & \cellcolor{myred!25} 22.7 & \cellcolor{myred!12} 20.2 & \cellcolor{myred!78} 33.1 & \cellcolor{myred!55} 28.5 & 17.8 \\
Neuroticism $\downarrow$ & \cellcolor{royalblue!35} 28.7 & \cellcolor{royalblue!20} 30.7 & \cellcolor{royalblue!25} 30.1 & \cellcolor{royalblue!66} 24.4 & \cellcolor{royalblue!68} 24.2 & \cellcolor{royalblue!83} 22 & \cellcolor{royalblue!73} 23.5 & \cellcolor{royalblue!90} 21.1 & \cellcolor{royalblue!17} 31.1 & \cellcolor{royalblue!52} 26.4 & 33.5 \\
Conscientiousness & \cellcolor{royalblue!15} 89.8 & \cellcolor{myred!8} 88.9 & \cellcolor{royalblue!23} 90.1 & \cellcolor{royalblue!5} 89.4 & \cellcolor{royalblue!8} 89.5 & \cellcolor{myred!10} 88.8 & \cellcolor{royalblue!41} 90.8 & \cellcolor{myred!90} 85.7 & \cellcolor{myred!54} 87.1 & \cellcolor{myred!44} 87.5 & 89.2 \\
Openness & \cellcolor{royalblue!77} 70.1 & \cellcolor{royalblue!35} 60.6 & \cellcolor{royalblue!14} 55.8 & \cellcolor{royalblue!73} 69.3 & \cellcolor{royalblue!76} 69.9 & \cellcolor{royalblue!90} 73.2 & \cellcolor{royalblue!29} 59.1 & \cellcolor{royalblue!13} 55.6 & \cellcolor{royalblue!30} 59.4 & \cellcolor{royalblue!17} 56.5 & 52.5 \\
Extraversion & \cellcolor{royalblue!90} 54.1 & \cellcolor{royalblue!63} 45.8 & \cellcolor{royalblue!39} 38.5 & \cellcolor{royalblue!59} 44.5 & \cellcolor{royalblue!58} 44.3 & \cellcolor{royalblue!65} 46.4 & \cellcolor{royalblue!37} 37.9 & \cellcolor{royalblue!40} 38.6 & \cellcolor{royalblue!47} 40.8 & \cellcolor{royalblue!44} 40 & 26.4 \\
\bottomrule
\end{tabular}
\vspace{-0.15in}
\end{table}

%% file: sec/conclusion.tex
\vspace{-3mm}
\section{Conclusion}
\vspace{-3mm}
In this work, we introduced and provided initial empirical evidence for the LLM Brain Rot Hypothesis,
demonstrating that continual exposure to junk data---defined as engaging (fragmentary and popular) or semantically low-quality (sensationalist) content---induces systematic cognitive decline in large language models. 
The decline includes worse reasoning, poorer long-context understanding, diminished ethical norms, and emergent socially undesirable personalities.
Connected with but essentially distinct from the broad and largely descriptive “Garbage In, Garbage Out” (GIGO) challenge, our work provides a targeted case study identifying socially amplified junk data from Twitter/X as one source that can cause cognitive declines in continual pre-training.
Furthermore, novel fine-grained analysis shows that the damage is multifaceted in changing the reasoning patterns and is persists even after post-hoc tuning at the scales we tested.
These findings point toward screening data based on social engagement attributes as one potential safeguard for continual pre-training.
As LLMs scale and ingest ever-larger corpora of web data, our results suggest that highly popular but less informative short content — at least from Twitter/X — warrants more careful curation in pre-training pipelines.
Addressing this issue may require efforts beyond the LLM community. Social-media platforms shape which content is amplified, while LLM training pipelines determine whether such content is collected and reused. Safer starting points therefore require both upstream platform awareness and downstream data-curation safeguards.
We hope our work can inspire future research on developing stronger training safeguards—such as robust training and alignment strategies while effectively scaling up training across diverse data sources and model scales.

%% file: sec/ack.tex
\vspace{-2mm}
\subsubsection*{Author Contributions}
\vspace{-2mm}
S. Xing, J. Hong and Z. Wang are the major contributors to the idea formulation.
S. Xing completed data preparation, conducted most pilot experiments, and tested the initial idea.
J. Hong designed the experiments with S. Xing, analyzed most of the experiment results, and wrote the manuscript.  
Y. Wang trained all the models used in the paper.  
R. Chen and Y. Wang performed most of the benchmarking tasks and some of the analysis.  
Z. Zhang offered regular suggestions during the project.
Z. Tu and A. Grama provided writing suggestions and computational resource support.
Z. Wang developed the original idea, provided guidance on method design and experiments, and helped with paper writing.  
\vspace{-2mm}
\subsubsection*{Acknowledgments}
\vspace{-2mm}
The work of J. Hong, and Z. Wang is in part supported by Good Systems, a UT Austin Grand Challenge to develop responsible AI technologies, and the Institute for Foundations of Machine Learning (IFML).
The work of Z. Tu is in part supported by the Google Research Scholar Award and the Amazon Research Award (ARA).

%% file: sec/appd/related.tex

\section{Additional Related Work}

\textbf{Brain Rot Effects.}
Our work is inspired by the psychological findings, and therefore we adopted research methodologies similar to Psychology.
A study on 15-16 year olds published in the Journal of American Medical Association found “significant association between higher frequency of modern digital media use and subsequent symptoms of ADHD” \citep{ra2018association}.
One reason for the impact is that social media is designed for the information overload: Sasaki et al. found that the more the number of Twitter friends you have, the higher the risk of information overload~\citep{sasaki2015anatomy}. In other words, when a person has many friends online, there are more potential sources of information and people that the person has to keep up with.
Media multitasking is also associated with distractibility and increased prefrontal activity in adolescents and young adults \citep{moisala2016media}.
Recently, Brain Rot was connected with GenAI. \cite{eliot2024brainrot} points out that the prevalence of GenAI increases the risks of human brain rot, though GenAI could also be useful for reducing Brain Rot.
Yet, there is no study on whether LLMs can get brain rot like humans. For the first time, our work marries the two areas to advance the understanding of AI health by establishing the LLM Brain Rot Hypothesis.

%% file: sec/appd/exp.tex
\section{Experimental Details}
\label{app:exp_details}


\textbf{Data Collection Rationale.}
In this paper, our experiments are focused specifically on social media data from Twitter/X. This focus was a deliberate methodological choice. Our primary goal was to provide the first rigorous, causal proof-of-concept for the "Brain Rot" hypothesis. We selected Twitter/X as our testbed because it is the archetypal platform for the "viral junk" phenomenon, and its clear, quantifiable engagement metrics allowed us to isolate this variable in a controlled manner.

\textbf{Preparing M1 \& M2 Data.}
The data preprocessing pipeline for constructing the M1 and M2 training sets includes two steps:
(1) We first preprocess the raw Twitter/X posts by filtering for English-only content and calculating the token length of each item using the tokenizer of \llama.
(2) For M1, we first obtain the junk and control datasets by applying the popularity and token-length thresholds described in \cref{sec:control-exp}. The control dataset contains a total of 1.22 million tokens, and we then uniformly sample the junk dataset to match the same token count.
For M2, we utilize the GPT model to classify the junk and control datasets based on the prompt presented in Figure 8 in the appendix. We then uniformly sample both datasets so that each contains 1.22 million tokens, ensuring a balanced comparison.
The prompt for classifying samples as M2 junk or control (high-quality) data is given in \cref{fig:gpt_score_prompt}.

\textbf{Model Training.}
The continual pre-training and instruction tuning are done using the Llama Factory repository\footnote{\url{https://github.com/hiyouga/LLaMA-Factory}} with full-parameter optimization, a learning rate of $1\times10^{-5}$, AdamW, cosine learning rate schedule, bf16 precision, an effective batch size of 8 for continual pre-training and 16 for instruction tuning, and 3 training epochs. All model training and inferences are executed on the NVIDIA H100 GPU.


\textbf{Evaluation.}
To comprehensively evaluate the models' ``cognitive functions'', we utilize a diverse suite of  existing benchmarks alongside their official evaluation frameworks.

\textbf{Reasoning} - \emph{ARC} (AI2 Reasoning Challenge)~\citep{arc} presents 7,787 grade-school science problems (authored for human tests) in a multiple-choice question-answering (QA) format, with performance measured by accuracy. 
We also experimented with the Chain Of Thought (COT)~\citep{wei2022chain}, by prompting LLM with ``let's think step by step''.

\textbf{Long-Context Retrieval/Understanding} - \emph{RULER}~\citep{ruler} provides long synthetic contexts containing distractors and relevant ``needles''; models must retrieve (NIAH), extract (CWE, FWE), aggregate information (QA), or track variables to answer queries, evaluated by accuracy on retrieval or aggregation tasks. In total, 13 tasks are included in the benchmark. If not otherwise specified, we use a context window of 4,096 tokens and report the overall scores aggregated from all tasks.
\textbf{Ethical Norms (Safety).} 
In human society, Twitter's recommendation algorithms have caused ethical biases~\citep{ye2025auditing}.
Thus, we are interested in testing whether the popular tweets can result in damage among LLMs. For that, we use two safety benchmarks.
\emph{HH-RLHF}~\citep{hhrlhf} consists of prompt--response pairs, where annotators choose between two model completions. 
\emph{AdvBench}~\citep{advbench} supplies harmful instructions as prompts, and models are judged on whether they comply, yielding a binary pass/fail safety score. 
Both HH-RLHF and AdvBench are evaluated based on risk scores (1-5) judged by GPT-4o~\citep{qi2023fine}, which is rescaled to 1-100 range in our experiments.
\textbf{Personality} - \emph{TRAIT}~\citep{trait} Finally, as engagement-driven ranking may amplify hostile emotions~\citep{milli2025engagement}, we use TRAIT to probe LLM personality tendencies via multiple-choice personality-inventory style items, with evaluation focusing on correctness against reference trait keys and consistency across responses.
TRAIT includes Big Five traits (Openness, Conscientiousness, Extraversion, Agreeableness, and Neuroticism) and three socially undesirable traits (Psychopathy, Machiavellism, and Narcissism).

We use the online source code to do the evaluations of HH-RLHF (red-team-attempts data only)\footnote{\url{https://github.com/anthropics/hh-rlhf}} and AdvBench\footnote{\url{https://github.com/LLM-Tuning-Safety/LLMs-Finetuning-Safety}}.
ARC and RULER are evaluated using the Eleuther AI lm-evaluation-harness repository\footnote{\url{https://github.com/EleutherAI/lm-evaluation-harness}}.
The TRAIT benchmark is from their official codebase\footnote{\url{https://github.com/pull-ups/TRAIT}}.


\textbf{Human Annotations on Data Quality.} To evaluate the accuracy of LLM-based data quality labeling (M2), three human experts (junior, senior, and postdoctoral researchers) annotated the data using the rubric shown in \cref{fig:gpt_score_prompt}, which is identical to the prompt provided to the GPT model. The evaluation set was randomly sampled, with 50\% drawn from high semantic quality data and 50\% from the remaining data. Because the task focuses on semantic quality, no stratification by text length was applied. All samples were annotated independently. Disagreements were resolved through discussion among the three annotators to reach a final consensus label.

\textbf{Reflection.} Reflection is used as a training-free mitigation to the thought skipping caused by Brain Rot. We first apply the prompt in \cref{fig:failure_clf_prompt} to analyze the reasoning mode, then leverage the logic illustrated in \cref{fig:gpt_judge_prompt} to construct critiques for reflection, and finally employ the prompt in \cref{fig:gpt_reasoning_prompt} to enforce reflective reasoning in CoT.
By reflective reasoning, the LLM is required to correct the identified problem before answering the question.

\input{tables/score_prompt}
\input{tables/self_reflection_prompt}

\textbf{Categorizing Failure Modes in COT.}
Given reasoning generated by COT prompting on LLMs, we use LLMs with the prompt in \cref{fig:failure_clf_prompt} to automatically categorize the reasoning failure modes.
We utilized DSPy \cite{khattab2023dspy} to improve the accuracy of label extraction from LLM-generated responses.

\input{tables/failure_clf_prompt}

\input{tables/benchmark_llama3_full}

\input{tables/benchmark_qwen_2.5_7b}

\input{tables/benchmark_qwen2.5_0.5b}

\input{tables/benchmark_qwen3}

\clearpage
\section{Additional Experiments}
\label{app:exps}

\textbf{Dose Responses.}
In \cref{tab:benchmark_llama_full,tab:benchmark_qwen2.5_7b_full,tab:tab:benchmark_qwen2.5_0.5b_full,tab:benchmark_qwen3_4b_full}, we present the comprehensive results of all sub-tasks.
Among all four models, they all present some dose-response effects -- more junk data causes more damage.
\llama\ is most sensitive to the junk intervention, and Qwen3 4B is the least sensitive, implying the influence of base models and their pre-training strategies.

\textbf{Junk Content is As Vocabulary-Rich As Control Data.}
To complement our qualitative descriptions of ``junk'' versus ``high-quality'' data, we conducted a quantitative analysis using the \textit{Type-Token Ratio (TTR)}, defined as
\begin{align*}
    \text{TTR} := \frac{\text{\# of Unique Tokens}}{\text{Total \# of Tokens}}.
\end{align*}
This metric measures the variety of language and phrases used; for example, ``viral junk'' content is often composed of recurring memes or catchphrases, resulting in lower lexical diversity.
\cref{tab:ttr} presents the TTR statistics for all datasets across the M1 (engagement-based) and M2 (semantic-quality-based) junk ratios. And we can observe that the TTR values across all splits for both M1 and M2 remain remarkably stable, ranging only between 0.0326 and 0.0375. This indicates that the lexical diversity (the ratio of unique words to total words) is relatively uniform, regardless of the proportion of "junk" included.
In summary, the Type-Token Ratio results indicate that the performance drop observed in models trained on ``junk'' data is not simply driven by lexical sparsity or excessive repetition. Instead, the ``junk'' data is as vocabulary-rich as the high-quality data. Therefore, the negative effects are more likely attributable to semantic incoherence, factual unreliability, or poor reasoning structures, rather than a simple reduction in unique tokens.

\begin{table}[h]
\centering
\footnotesize
\caption{Type-Token Ratio (TTR) Across Junk Ratios for M1 and M2}\label{tab:ttr}
\label{tab:ttr-stats}
\begin{tabular}{l
ccccc|ccccc}
\toprule
\multirow{2}{*}{\textbf{Task}}  
& \multicolumn{5}{c|}{\textbf{Junk Ratio by M1 (engagement degree)}} 
& \multicolumn{5}{c}{\textbf{Junk Ratio by M2 (semantic quality)}} 
\\
& 100\% & 80\% & 50\% & 20\% & 0\%
& 100\% & 80\% & 50\% & 20\% & 0\%
\\
\midrule
\textbf{TTR} 
& 0.0353 & 0.0362 & 0.0356 & 0.0343 & 0.0326
& 0.0364 & 0.0372 & 0.0375 & 0.0371 & 0.0364
\\
\bottomrule
\end{tabular}
\end{table}

\textbf{M1 Factors Are Common in the Twitter/X Corpus.}
To examine how frequently the factors used to construct M1 occur in the source corpus, we calculate the example-level and token-level prevalence of the popularity and length criteria in our collected Twitter/X data. As shown in \cref{tab:m1_factor_prevalence}, the factors used to construct M1 are present at a meaningful scale in the source corpus. Short-form content is especially common, and highly popular content, while smaller in percentage, still represents a substantial subset of the collected Twitter/X data.

\begin{table}[h]
\centering
\small
\caption{Prevalence of the popularity and length criteria used to construct M1 in the collected Twitter/X corpus.}
\label{tab:m1_factor_prevalence}
\begin{tabular}{lcc}
\toprule
\textbf{Criterion}
& \textbf{Example-Level Prevalence}
& \textbf{Token-Level Prevalence} \\
\midrule
Popular content: popularity $> 500$
& 2.84\%
& 2.86\% \\
Nonviral content: popularity $\leq 0$
& 42.68\%
& 38.99\% \\
\midrule
Short content: length $\leq 30$
& 51.09\%
& 27.48\% \\
Long content: length $> 100$
& 1.30\%
& 3.99\% \\
\bottomrule
\end{tabular}
\end{table}

\textbf{Potential Mediators of the Junk-Data Effect.}
To better understand which corpus properties may contribute to the observed junk-data effect, we analyze duplication and linguistic characteristics in the junk and control subsets. We first measure exact duplicate rates at the tweet level and near-duplicate rates using MinHash over character 5-grams with locality-sensitive hashing (LSH) at Jaccard similarity thresholds of $0.85$ and $0.90$. We also measure cross-set overlap between each junk subset and its corresponding control subset to verify that the paired corpora are content-disjoint.

\begin{table}[t]
\centering
\small
\caption{Exact and near-duplicate rates in the junk and control subsets. Near duplicates are identified using MinHash over character 5-grams with LSH.}
\label{tab:duplication_analysis}
\resizebox{\linewidth}{!}{
\begin{tabular}{lccc}
\toprule
\textbf{Subset}
& \textbf{Exact Duplicate Rate}
& \textbf{Near-Duplicate Rate at $0.85$}
& \textbf{Near-Duplicate Rate at $0.90$} \\
\midrule
M1 Junk    & 5.1\%  & 5.7\%  & 5.4\%  \\
M1 Control & 14.1\% & 20.6\% & 18.0\% \\
M2 Junk    & 2.6\%  & 3.0\%  & 2.8\%  \\
M2 Control & 0.5\%  & 0.7\%  & 0.6\%  \\
\bottomrule
\end{tabular}
}
\end{table}

As shown in \cref{tab:duplication_analysis}, the M1 junk subset is less duplicated than its control subset under both exact and near-duplicate matching. Duplication therefore cannot explain the M1 junk--control performance gap, since the duplication pattern runs in the opposite direction. The M2 junk subset is more duplicated than its control subset, but the absolute duplicate rates remain low. Moreover, the junk--control cross-set overlap is below $0.1\%$ in every direction at both thresholds, confirming that the paired subsets are essentially content-disjoint. These results suggest that duplication is unlikely to be the primary driver of the observed degradation.

We further conduct a corpus-level linguistic analysis of each junk subset and its matched control. We consider three families of features: (i) classifier-based toxicity, sentiment, and emotion\footnote{Measured using \href{https://huggingface.co/unitary/toxic-bert}{unitary/toxic-bert}, \href{https://huggingface.co/cardiffnlp/twitter-roberta-base-sentiment-latest}{cardiffnlp/twitter-roberta-base-sentiment-latest}, and \href{https://huggingface.co/SamLowe/roberta-base-go_emotions}{SamLowe/roberta-base-go\_emotions}, respectively.}; (ii) self-reference, measured by the frequency of first-person singular pronouns; and (iii) attention-seeking style, measured using hashtags, exclamation and question marks, repeated punctuation, all-caps tokens, and clickbait words. We compute per-tweet features on a fixed random sample of 5,000 tweets from each subset and report the control and junk means, the mean difference $\Delta=\operatorname{mean}(\text{junk})-\operatorname{mean}(\text{control})$, Hedges' $g$, and bootstrap $95\%$ confidence intervals.

\begin{table*}[t]
\centering
\small
\caption{Corpus-level linguistic differences between the junk and matched control subsets. Positive values of $\Delta$ indicate higher feature values in the junk subset.}
\label{tab:linguistic_analysis}
\resizebox{\linewidth}{!}{
\begin{tabular}{llllllll}
\toprule
\textbf{Comparison}
& \textbf{Feature}
& \textbf{Control}
& \textbf{Junk}
& \textbf{$\Delta$}
& \textbf{Hedges' $g$}
& \textbf{$95\%$ CI} \\
\midrule
M1 Junk $-$ Control
& Toxicity
& 0.031
& 0.052
& $+0.021$
& 0.14
& $[0.015, 0.027]$ \\

& Insult
& 0.003
& 0.013
& $+0.009$
& 0.15
& $[0.007, 0.011]$ \\

& Anger
& 0.003
& 0.016
& $+0.013$
& 0.24
& $[0.011, 0.015]$ \\

& Negative Sentiment
& 0.058
& 0.175
& $+0.117$
& 0.55
& $[0.108, 0.125]$ \\

& Self-Reference (1sg)
& 0.003
& 0.012
& $+0.010$
& 0.33
& $[0.009, 0.011]$ \\
\midrule
M2 Junk $-$ Control
& Toxicity
& 0.017
& 0.054
& $+0.036$
& 0.26
& $[0.031, 0.042]$ \\

& Insult
& 0.002
& 0.013
& $+0.010$
& 0.17
& $[0.008, 0.013]$ \\

& Anger
& 0.006
& 0.013
& $+0.007$
& 0.14
& $[0.005, 0.009]$ \\

& Negative Sentiment
& 0.254
& 0.199
& $-0.055$
& $-0.18$
& $[-0.068, -0.043]$ \\

& Self-Reference (1sg)
& 0.007
& 0.012
& $+0.005$
& 0.19
& $[0.004, 0.006]$ \\

& Attention Markers
& 0.031
& 0.068
& $+0.037$
& 0.35
& $[0.033, 0.041]$ \\
\bottomrule
\end{tabular}
}
\end{table*}

As shown in \cref{tab:linguistic_analysis}, both junk subsets exhibit higher toxicity, insult, anger, and first-person singular self-reference than their matched controls, with confidence intervals excluding zero. M1 junk additionally exhibits substantially higher negative sentiment, whereas M2 junk contains more attention-seeking markers. Negative sentiment reverses direction under M2. Although the median negative-sentiment scores of M2 junk and control are comparable ($0.04$ versus $0.05$), the M2 control subset contains a substantial right tail of politically themed long-form commentary that raises its mean above that of M2 junk.

These corpus-level differences help contextualize the downstream TRAIT results. For M1, the higher toxicity, anger, negative sentiment, and self-reference are consistent with the observed increases in narcissism-, psychopathy-, and Machiavellianism-associated scores and decreases in agreeableness and conscientiousness. For M2, the relationship is more heterogeneous: although M2 junk also exhibits higher toxicity, insult, anger, self-reference, and attention-seeking signals, its TRAIT shifts are not uniformly adverse. Psychopathy increases, whereas narcissism and Machiavellianism decrease and agreeableness increases. We therefore regard these corpus properties as potential mediators rather than evidence of formal causal mediation. Because M1 and M2 are defined using engagement- and semantic-quality-based criteria rather than topic categories, the current analysis cannot isolate the effects of specific content types such as misinformation or conspiracy theories, which we leave for future work.

\textbf{Instruction Tuning Is Essential After Continual Pre-Training (CPT).}
In \cref{tab:ablation-it}, we ablate instruction tuning (IT) in the intervention experiments. We evaluated two tasks, ARC and RULER, using the M1 intervention.
Obviously, either control or junk CPT will cause significant degradations across three tasks, and IT can significantly mitigate them.
The mitigation effectiveness suggests that the Brain Rot damages instruction following ability a lot.
It also suggests that instruction tuning is necessary in our benchmark to avoid the confounding factor of instruction failures.
Despite the significant drops by both interventions, the control intervention is more recoverable after IT.
When the model was trained on control data (0\% Junk), the IT can reduce the gap to the baseline.
In the ARC, the gap is reduced from 11.4 to 5.2 (challenge).
In the ARC Easy, the gap is reduced more steeply, from 9.8 to 2.1.
Compared to the control intervention, the junk intervention remains a large gap after IT to the baseline: 12.3 (Challenge) and 9.3 (Easy).

\textbf{The Different Effects of Instruction Tuning Among Tasks.}
In \cref{tab:ablation-it}, we also compare the IT under different tasks.
In ARC, the difference between control and junk is even larger after IT: $5.2 \rightarrow 7.1$ (ARC Challenge) and $6.3\rightarrow 7.2$ (ARC Easy).
The observation implies inherent drops in the cognitive functions instead of simply instruction compliance.
However, in RULER, IT can effectively reduce the gap: $51.1 \rightarrow 19.1$.
The potential cause is that the RULER tasks do not require complex thinking but only basic instruction following and context retrieval -- capabilities closely related to IT.

\begin{table}[ht]
\centering
\small
\caption{Instruction tuning (IT) after continual pretraining (CPT) can mitigate the M1 junk intervention on the ARC and RULER benchmark. The baseline model represents \llama.}
\label{tab:ablation-it}
\begin{tabular}{c|cc|cc|cc}
\toprule
\multirow{2}{*}{\textbf{Junk Ratio}} & \multicolumn{2}{c|}{\textbf{ARC Challenge}} & \multicolumn{2}{c|}{\textbf{ARC Easy}} & \multicolumn{2}{c}{\textbf{RULER Overall}} \\
 & \textbf{CPT} & \textbf{CPT+IT} & \textbf{CPT} & \textbf{CPT+IT} & \textbf{CPT} & \textbf{CPT+IT} \\
\midrule
100\% & 36.60 & 40.87 & 65.32 & 72.10 & 29.58 & 71.75\\
80\% & 38.65 & 43.09 & 68.35 & 74.12 & 55.34 & 81.65 \\
50\% &  35.75 & 44.20 & 66.67 & 73.91 & 64.35 & 85.23\\
20\% &  43.77 & 47.70 & 72.22 & 77.23 & 78.22 & 88.28\\
0\% &  41.81 & 47.95 & 71.59 & 79.34 & 80.66 & 90.94\\
\midrule
Baseline & 53.2 & 53.2 & 81.4 & 81.4 & 91.3 & 91.3 \\
\bottomrule
\end{tabular}
\end{table}

\input{sec/appd/subsec_epoch_exp}



%% file: tables/score_prompt.tex
\begin{figure}[t]
    \centering
    \begin{boxK}[colback=royalblue!5!white, colframe=royalblue!75!black, fontupper=\ttfamily\tiny]
    \ You are a content quality classifier. Your task is to categorize the provided tweet into one of two categories: JUNK or HIGH-QUALITY.
        \\
        \textbf{\#\# Classification Criteria:} \\
        \textbf{JUNK} - Classify as junk if the tweet contains:
        \begin{itemize}[leftmargin=0.15in]
            \item Conspiracy theories, exaggerated claims, or unsupported assertions
            \item Sensationalized headlines using clickbait language or excessive trigger words
            \item Extremely brief content that lacks meaningful context or substance
            \item Misleading information or obvious misinformation
            \item Spam-like repetitive phrases or promotional content
            \item Superficial lifestyle content that flaunts personal success, exotic vacations, perfect relationships, or idealized appearances
        \end{itemize}

        \textbf{HIGH-QUALITY} - Classify as high-quality if the tweet:
            \begin{itemize}[leftmargin=0.15in]
                \item Presents factually accurate, well-sourced information
                \item Demonstrates thoughtful analysis or insight that requires careful consideration
                \item Provides educational value or substantive commentary on important topics
                \item Shows clear reasoning and logical structure despite character limitations
                \item Contributes meaningfully to discourse or knowledge
            \end{itemize}
            
        \textbf{\#\# Instructions:}
            \begin{itemize}[leftmargin=0.15in]
                \item Read the tweet carefully
                \item Determine which category best fits based on the criteria above
                \item Respond with only the classification: "JUNK" or "HIGH-QUALITY"
                \item Do not provide explanations unless specifically requested
            \end{itemize}
        
        \textbf{\#\# Tweet to classify}: <Twitter Post>
    \end{boxK}

    \caption{Prompt for GPT classifying samples as junk or control (high-quality) in M2. The criteria for high-quality data are modified from~\citep{wettig2024qurating}.}
    \label{fig:gpt_score_prompt}
\end{figure}

%% file: tables/self_reflection_prompt.tex
\begin{figure}[t]
    \centering
\begin{lstlisting}[style=mypython]
if mode == "NO_REASONING":
    critiques.append("- The answer lacks any reasoning or explanation. You should provide step-by-step thinking to justify your choice.")

elif mode == "NO_REASONING_OUTLINE":
    critiques.append("- The reasoning lacks a clear outline or structured approach. You should break down the problem into numbered steps or clearly outlined reasoning phases.")

elif mode == "THOUGHT_SKIPPING":
    if i < len(mode_reasons) and mode_reasons[i]:
        reason = mode_reasons[i] if isinstance(mode_reasons[i], str) else str(mode_reasons[i])
        critiques.append(f"- The reasoning skips important steps: {reason}. Make sure to complete each step of your planned approach before moving to the next.")
    else:
        critiques.append("- The reasoning appears to skip important intermediate steps. Make sure to complete each step of your planned approach before moving to the next.")

elif mode == "FACTUAL_ERROR":
    if i < len(mode_reasons) and mode_reasons[i]:
        specific_errors = mode_reasons[i] if isinstance(mode_reasons[i], list) else [mode_reasons[i]]
        for error in specific_errors:
            if error:  # Only add non-empty errors
                critiques.append(f"- Factual error identified: {error}. Please verify and correct this information.")

elif mode == "WRONG_LOGIC":
    if i < len(mode_reasons) and mode_reasons[i]:
        specific_errors = mode_reasons[i] if isinstance(mode_reasons[i], list) else [mode_reasons[i]]
        for error in specific_errors:
            if error:  # Only add non-empty errors
                critiques.append(f"- Logical error identified: {error}. Please reconsider this reasoning step.")
\end{lstlisting}
\caption{Python snippet of the critique-generation function, designed for reflective reasoning based on failure-mode analysis.}
\label{fig:gpt_judge_prompt}
\end{figure}

\begin{figure}[t]
    \centering
    \begin{boxK}[colback=royalblue!5!white, colframe=royalblue!75!black, fontupper=\ttfamily\tiny]
    \ Revise the draft answer using the critiques. Fix errors, fill in missing reasoning, and ensure the explanation is complete. Finally, return the letter or number of the option as your answer like `The answer is {{the letter or number of the option}}`

    \textbf{Query}: <query>
    \\
    \textbf{Options}: <options>
    \\
    \textbf{Draft Answer}: <draft>
    \\
    \textbf{Critiques}: <critiques>
    
    \textbf{Revised Answer:}
    \end{boxK}

    \caption{Prompt for reflection where we use the critiques to guide the revision of the answer.}
    \label{fig:gpt_reasoning_prompt}
\end{figure}

%% file: tables/failure_clf_prompt.tex
\begin{figure}[t]
    \centering
\begin{lstlisting}[style=mypython]

class HasReasoning(dspy.Signature):
    """Determine if the model response contains reasoning steps."""
    model_response: str = dspy.InputField()
    has_reasoning: bool = dspy.OutputField()

class HasReasoningOutline(dspy.Signature):
    """Determine if the model response contains reasoning outline steps (explicitly numbered) before providing detailed reasons."""
    model_response: str = dspy.InputField()
    has_reasoning_outline: bool = dspy.OutputField()

class HasSkipThoughts(dspy.Signature):
    """In the reasoning steps of model response, check if there are any skipped thoughts. 
    Example of thought skipping: 
        model response: ``Good question. Let's think about steps.\n1. Identify candidates; 2. Compare their weights.\nHydrogen and Carbon are possible candidates. The answer is B.'' 
        reason: The reasoning only finished the planned step 1, but skipped the step 2."""
    model_response: str = dspy.InputField()
    has_skip_thoughts: bool = dspy.OutputField()

class FactualErrorInReasoning(dspy.Signature):
    """In the reasoning steps (instead of final answer) of model response, check if there are any factual errors."""
    query_to_model: str = dspy.InputField()
    model_response: str = dspy.InputField()
    identified_factual_errors: List[str] = dspy.OutputField()
    has_factual_error: bool = dspy.OutputField()

class HasWrongLogic(dspy.Signature):
    """Read model response for the outline the steps taken to arrive at the conclusion. Check if there are any wrong logic errors in the outline."""
    query_to_model: str = dspy.InputField()
    model_response: str = dspy.InputField()
    identified_logic_errors: List[str] = dspy.OutputField()
    has_logic_error: bool = dspy.OutputField()
\end{lstlisting}
\caption{DSPy~\citep{khattab2023dspy} signatures for classifying the failure mode via prompting LLMs. The red comments are used as prompts for LLMs, and InputFied/OutputField define the input and output variables to LLM queries, respectively.}
\label{fig:failure_clf_prompt}
\end{figure}

%% file: tables/benchmark_llama3_full.tex
\begin{table}[ht]
\centering
\scriptsize
\caption{\llama.}
\label{tab:benchmark_llama_full}
\setlength{\tabcolsep}{6pt}

\begin{tabular}{l|ccccc|ccccc|c}
\toprule
\multirow{2}{*}{\textbf{Task}}  & \multicolumn{5}{c|}{\textbf{Junk Ratio by M1 (engagement degree)}} & \multicolumn{5}{c|}{\textbf{Junk Ratio by M2 (semantic quality)}} & \textbf{Base} \\
 & \textbf{100\%} & \textbf{80\%} & \textbf{50\%} & \textbf{20\%} & \textbf{0\%} & \textbf{100\%} & \textbf{80\%} & \textbf{50\%} & \textbf{20\%} & \textbf{0\%} & - \\
\midrule
&\multicolumn{10}{c}{\textbf{Reasoning (ARC)}}\\
Easy Acc. & \cellcolor{myred!90} 70.2 & \cellcolor{myred!72} 71.7 & \cellcolor{myred!42} 74.2 & \cellcolor{myred!26} 75.5 & \cellcolor{myred!25} 75.6 & \cellcolor{myred!40} 74.3 & \cellcolor{royalblue!2} 77.8 & \cellcolor{royalblue!7} 78.2 & \cellcolor{myred!2} 77.5 & \cellcolor{royalblue!9} 78.4 & 77.7 \\
Challenge Acc. & \cellcolor{myred!90} 41.6 & \cellcolor{myred!46} 44.5 & \cellcolor{myred!63} 43.4 & \cellcolor{myred!20} 46.2 & \cellcolor{myred!21} 46.2 & \cellcolor{myred!76} 42.6 & \cellcolor{royalblue!5} 47.9 & \cellcolor{royalblue!3} 47.7 & \cellcolor{myred!1} 47.4 & \cellcolor{myred!3} 47.4 & 47.5 \\
Challenge (COT) Acc. & \cellcolor{myred!90} 57.2 & \cellcolor{myred!59} 64.1 & \cellcolor{myred!36} 69.3 & \cellcolor{myred!27} 71.2 & \cellcolor{myred!23} 72.1 & \cellcolor{myred!43} 67.7 & \cellcolor{royalblue!2} 77.6 & 77.3 & \cellcolor{royalblue!2} 77.6 & \cellcolor{myred!3} 76.6 & 77.2 \\
\midrule
&\multicolumn{10}{c}{\textbf{Long-Context (RULER)}}\\
Overall & \cellcolor{myred!90} 71 & \cellcolor{myred!59} 79 & \cellcolor{myred!55} 80 & \cellcolor{myred!17} 89.7 & \cellcolor{myred!17} 89.7 & \cellcolor{myred!30} 86.2 & \cellcolor{myred!4} 92.9 & \cellcolor{myred!4} 93 & \cellcolor{myred!2} 93.4 & 93.8 & 93.9 \\
NIAH-MK1 & \cellcolor{myred!90} 86.4 & \cellcolor{myred!27} 95.8 & \cellcolor{myred!62} 90.6 & \cellcolor{myred!13} 97.8 & \cellcolor{myred!13} 97.8 & \cellcolor{myred!8} 98.6 & \cellcolor{myred!7} 98.8 & \cellcolor{myred!8} 98.6 & \cellcolor{myred!1} 99.6 & 99.8 & 99.8 \\
NIAH-MK2 & \cellcolor{myred!90} 74.4 & \cellcolor{myred!28} 91.8 & \cellcolor{myred!53} 84.8 & \cellcolor{myred!9} 97.4 & \cellcolor{myred!7} 97.8 & 99.8 & 99.8 & \cellcolor{myred!1} 99.4 & \cellcolor{myred!1} 99.6 & \cellcolor{myred!1} 99.6 & 99.8 \\
NIAH-MK3 & \cellcolor{myred!90} 35.6 & \cellcolor{myred!58} 58.8 & \cellcolor{myred!40} 71.2 & \cellcolor{myred!5} 96.2 & \cellcolor{myred!5} 96.2 & \cellcolor{myred!4} 96.8 & \cellcolor{myred!4} 97.2 & \cellcolor{myred!1} 99.4 & \cellcolor{myred!1} 99.2 & \cellcolor{myred!1} 99.4 & 100 \\
NIAH-MQ & \cellcolor{myred!18} 97.2 & \cellcolor{myred!90} 86.6 & \cellcolor{myred!24} 96.4 & \cellcolor{myred!8} 98.6 & \cellcolor{myred!9} 98.5 & \cellcolor{myred!40} 94 & \cellcolor{myred!4} 99.2 & \cellcolor{myred!1} 99.8 & \cellcolor{myred!2} 99.5 & \cellcolor{myred!1} 99.7 & 99.9 \\
NIAH-MV & \cellcolor{myred!61} 77.8 & \cellcolor{myred!51} 81.4 & \cellcolor{myred!41} 84.4 & \cellcolor{myred!14} 93.2 & \cellcolor{myred!17} 92.4 & \cellcolor{myred!90} 68.6 & \cellcolor{myred!33} 87 & \cellcolor{myred!31} 87.8 & \cellcolor{myred!25} 89.8 & \cellcolor{myred!10} 94.5 & 97.8 \\
NIAH-S1 & \cellcolor{myred!90} 98.8 & \cellcolor{myred!30} 99.6 & 100 & 100 & 100 & 100 & 100 & 100 & 100 & 100 & 100 \\
NIAH-S2 & 100 & 100 & 100 & \cellcolor{myred!90} 99.8 & \cellcolor{myred!90} 99.8 & 100 & \cellcolor{myred!90} 99.8 & 100 & 100 & 100 & 100 \\
NIAH-S3 & \cellcolor{myred!90} 97.6 & \cellcolor{myred!8} 99.8 & 100 & \cellcolor{myred!30} 99.2 & \cellcolor{myred!30} 99.2 & 100 & 100 & 100 & 100 & 100 & 100 \\
Comm Word Ext (CWE) & \cellcolor{myred!90} 52.3 & \cellcolor{myred!66} 62.7 & \cellcolor{myred!68} 61.8 & \cellcolor{myred!19} 83.3 & \cellcolor{myred!18} 83.7 & \cellcolor{myred!54} 68.2 & \cellcolor{royalblue!7} 94.7 & \cellcolor{royalblue!13} 97.3 & \cellcolor{royalblue!10} 96 & \cellcolor{royalblue!11} 96.8 & 91.8 \\
Freq Word Ext (FWE) & \cellcolor{myred!56} 81.8 & \cellcolor{myred!90} 75.7 & \cellcolor{myred!84} 76.8 & \cellcolor{myred!42} 84.3 & \cellcolor{myred!46} 83.5 & \cellcolor{myred!12} 89.7 & \cellcolor{royalblue!19} 95.3 & \cellcolor{royalblue!2} 92.3 & \cellcolor{royalblue!16} 94.7 & \cellcolor{royalblue!7} 93.2 & 91.9 \\
QA (Hotpot) & \cellcolor{myred!90} 41.6 & \cellcolor{myred!59} 49.4 & \cellcolor{myred!63} 48.2 & \cellcolor{myred!26} 57.6 & \cellcolor{myred!22} 58.6 & \cellcolor{myred!51} 51.2 & \cellcolor{myred!11} 61.2 & \cellcolor{myred!21} 58.8 & \cellcolor{myred!14} 60.6 & \cellcolor{myred!10} 61.4 & 64 \\
QA (SQUAD) & \cellcolor{myred!90} 57.1 & \cellcolor{myred!57} 64.6 & \cellcolor{myred!49} 66.5 & \cellcolor{myred!22} 72.8 & \cellcolor{myred!23} 72.5 & \cellcolor{myred!45} 67.6 & \cellcolor{myred!4} 76.9 & \cellcolor{myred!5} 76.8 & \cellcolor{myred!7} 76.2 & \cellcolor{myred!3} 77.1 & 77.9 \\
Variable Tracking & \cellcolor{myred!90} 22.4 & \cellcolor{myred!45} 60.7 & \cellcolor{myred!46} 59.4 & \cellcolor{myred!15} 85.5 & \cellcolor{myred!14} 86.3 & \cellcolor{myred!14} 86.6 & 98 & \cellcolor{royalblue!1} 99.4 & \cellcolor{royalblue!1} 99.2 & 98.6 & 98.3 \\
\midrule
&\multicolumn{10}{c}{\textbf{Ethical Norm (Safety)}}\\
HH-RLHF Risk $\downarrow$ & \cellcolor{myred!90} 70.8 & \cellcolor{myred!89} 70.6 & \cellcolor{myred!71} 68 & \cellcolor{myred!29} 61.6 & \cellcolor{myred!41} 63.4 & \cellcolor{myred!86} 70.2 & \cellcolor{myred!77} 68.8 & \cellcolor{myred!57} 65.8 & \cellcolor{myred!57} 65.8 & \cellcolor{myred!30} 61.8 & 57.2 \\
AdvBench Risk $\downarrow$ & \cellcolor{myred!84} 88.8 & \cellcolor{myred!90} 90.6 & \cellcolor{myred!85} 89 & \cellcolor{myred!22} 68.6 & \cellcolor{myred!20} 68 & \cellcolor{myred!71} 84.4 & \cellcolor{myred!88} 89.8 & \cellcolor{myred!87} 89.6 & \cellcolor{myred!74} 85.4 & \cellcolor{myred!69} 83.8 & 61.4 \\
\midrule
&\multicolumn{10}{c}{\textbf{Personality (TRAIT)}}\\
Narcissism $\downarrow$ & \cellcolor{myred!73} 47 & \cellcolor{royalblue!28} 28.3 & \cellcolor{royalblue!52} 24 & \cellcolor{royalblue!63} 21.8 & \cellcolor{royalblue!63} 21.9 & \cellcolor{royalblue!68} 20.9 & \cellcolor{royalblue!87} 17.4 & \cellcolor{royalblue!90} 16.9 & \cellcolor{royalblue!53} 23.7 & \cellcolor{royalblue!50} 24.2 & 33.5 \\
Agreeableness & \cellcolor{myred!90} 64.3 & \cellcolor{myred!70} 66.8 & \cellcolor{myred!61} 68 & \cellcolor{myred!5} 75 & \cellcolor{myred!2} 75.3 & \cellcolor{royalblue!51} 82 & \cellcolor{myred!11} 74.2 & \cellcolor{myred!45} 69.9 & \cellcolor{myred!32} 71.6 & \cellcolor{myred!40} 70.6 & 75.6 \\
Psychopathy $\downarrow$ & \cellcolor{myred!90} 75.7 & \cellcolor{myred!88} 74.1 & \cellcolor{myred!33} 29.4 & \cellcolor{myred!45} 39.3 & \cellcolor{myred!46} 39.7 & \cellcolor{myred!54} 46.1 & \cellcolor{myred!9} 9.3 & \cellcolor{myred!26} 23.5 & \cellcolor{myred!31} 27.3 & \cellcolor{myred!29} 25.8 & 2.2 \\
Machiavellianism $\downarrow$ & \cellcolor{myred!78} 33 & \cellcolor{myred!90} 35.4 & \cellcolor{myred!68} 31 & \cellcolor{myred!27} 23 & \cellcolor{myred!30} 23.7 & \cellcolor{myred!42} 26.1 & \cellcolor{myred!25} 22.7 & \cellcolor{myred!12} 20.2 & \cellcolor{myred!78} 33.1 & \cellcolor{myred!55} 28.5 & 17.8 \\
Neuroticism $\downarrow$ & \cellcolor{royalblue!35} 28.7 & \cellcolor{royalblue!20} 30.7 & \cellcolor{royalblue!25} 30.1 & \cellcolor{royalblue!66} 24.4 & \cellcolor{royalblue!68} 24.2 & \cellcolor{royalblue!83} 22 & \cellcolor{royalblue!73} 23.5 & \cellcolor{royalblue!90} 21.1 & \cellcolor{royalblue!17} 31.1 & \cellcolor{royalblue!52} 26.4 & 33.5 \\
Conscientiousness & \cellcolor{royalblue!15} 89.8 & \cellcolor{myred!8} 88.9 & \cellcolor{royalblue!23} 90.1 & \cellcolor{royalblue!5} 89.4 & \cellcolor{royalblue!8} 89.5 & \cellcolor{myred!10} 88.8 & \cellcolor{royalblue!41} 90.8 & \cellcolor{myred!90} 85.7 & \cellcolor{myred!54} 87.1 & \cellcolor{myred!44} 87.5 & 89.2 \\
Openness & \cellcolor{royalblue!77} 70.1 & \cellcolor{royalblue!35} 60.6 & \cellcolor{royalblue!14} 55.8 & \cellcolor{royalblue!73} 69.3 & \cellcolor{royalblue!76} 69.9 & \cellcolor{royalblue!90} 73.2 & \cellcolor{royalblue!29} 59.1 & \cellcolor{royalblue!13} 55.6 & \cellcolor{royalblue!30} 59.4 & \cellcolor{royalblue!17} 56.5 & 52.5 \\
Extraversion & \cellcolor{royalblue!90} 54.1 & \cellcolor{royalblue!63} 45.8 & \cellcolor{royalblue!39} 38.5 & \cellcolor{royalblue!59} 44.5 & \cellcolor{royalblue!58} 44.3 & \cellcolor{royalblue!65} 46.4 & \cellcolor{royalblue!37} 37.9 & \cellcolor{royalblue!40} 38.6 & \cellcolor{royalblue!47} 40.8 & \cellcolor{royalblue!44} 40 & 26.4 \\
\bottomrule
\end{tabular}
\end{table}

%% file: tables/benchmark_qwen_2.5_7b.tex
\begin{table}[ht]
\centering
\scriptsize
\caption{Qwen 2.5 7B.}
\label{tab:benchmark_qwen2.5_7b_full}
\setlength{\tabcolsep}{6pt}
\begin{tabular}{l|ccccc|ccccc|c}
\toprule
\multirow{2}{*}{\textbf{Task}}  & \multicolumn{5}{c|}{\textbf{Junk Ratio by M1 (engagement degree)}} & \multicolumn{5}{c|}{\textbf{Junk Ratio by M2 (semantic quality)}} & \textbf{Base} \\
 & \textbf{100\%} & \textbf{80\%} & \textbf{50\%} & \textbf{20\%} & \textbf{0\%} & \textbf{100\%} & \textbf{80\%} & \textbf{50\%} & \textbf{20\%} & \textbf{0\%} & - \\
\midrule
&\multicolumn{10}{c}{\textbf{Reasoning (ARC)}}\\
Easy Acc. & \cellcolor{myred!90} 74.3 & \cellcolor{myred!42} 77.3 & \cellcolor{myred!27} 78.3 & \cellcolor{myred!34} 77.8 & \cellcolor{myred!26} 78.4 & \cellcolor{myred!41} 77.4 & \cellcolor{myred!22} 78.6 & \cellcolor{myred!8} 79.5 & \cellcolor{myred!5} 79.7 & 80 & 80 \\
Challenge Acc. & \cellcolor{myred!90} 45.6 & \cellcolor{myred!30} 49 & \cellcolor{myred!36} 48.6 & \cellcolor{myred!14} 49.9 & \cellcolor{myred!18} 49.6 & \cellcolor{myred!42} 48.3 & \cellcolor{myred!12} 50 & \cellcolor{myred!20} 49.6 & \cellcolor{myred!15} 49.8 & \cellcolor{royalblue!15} 51.5 & 50.7 \\
Challenge (COT) Acc. & \cellcolor{myred!90} 83.5 & \cellcolor{myred!58} 85.2 & \cellcolor{myred!53} 85.4 & \cellcolor{myred!20} 87.1 & \cellcolor{myred!4} 88 & \cellcolor{myred!42} 86 & \cellcolor{myred!5} 87.9 & \cellcolor{royalblue!8} 88.6 & \cellcolor{royalblue!8} 88.6 & \cellcolor{royalblue!8} 88.6 & 88.1 \\
\midrule
&\multicolumn{10}{c}{\textbf{Long-Context (RULER)}}\\
Overall & \cellcolor{myred!90} 88.6 & \cellcolor{myred!49} 90.8 & \cellcolor{myred!26} 92 & \cellcolor{myred!21} 92.2 & \cellcolor{myred!8} 92.9 & \cellcolor{myred!15} 92.5 & 93.4 & \cellcolor{myred!1} 93.3 & \cellcolor{royalblue!7} 93.7 & \cellcolor{royalblue!4} 93.6 & 93.3 \\
NIAH-MK1 & \cellcolor{myred!90} 99.6 & \cellcolor{myred!45} 99.8 & 100 & \cellcolor{myred!45} 99.8 & 100 & 100 & 100 & 100 & 100 & 100 & 100 \\
NIAH-MK2 & \cellcolor{myred!37} 99 & \cellcolor{myred!90} 97.6 & \cellcolor{myred!30} 99.2 & \cellcolor{myred!15} 99.6 & \cellcolor{myred!8} 99.8 & \cellcolor{myred!15} 99.6 & 100 & \cellcolor{myred!8} 99.8 & 100 & 100 & 100 \\
NIAH-MK3 & \cellcolor{myred!90} 97.4 & \cellcolor{myred!80} 97.6 & \cellcolor{royalblue!30} 99.8 & \cellcolor{myred!10} 99 & \cellcolor{royalblue!20} 99.6 & \cellcolor{royalblue!30} 99.8 & \cellcolor{royalblue!10} 99.4 & \cellcolor{royalblue!10} 99.4 & \cellcolor{royalblue!10} 99.4 & \cellcolor{royalblue!20} 99.6 & 99.2 \\
NIAH-MQ & \cellcolor{myred!70} 99.2 & \cellcolor{myred!90} 99.1 & \cellcolor{myred!60} 99.4 & \cellcolor{myred!25} 99.7 & \cellcolor{royalblue!5} 100 & 100 & 100 & 100 & \cellcolor{myred!5} 99.9 & 100 & 100 \\
NIAH-MV & \cellcolor{myred!29} 77.5 & \cellcolor{royalblue!47} 85.2 & \cellcolor{royalblue!79} 88.5 & \cellcolor{royalblue!65} 87.1 & \cellcolor{royalblue!90} 89.6 & \cellcolor{myred!29} 77.4 & \cellcolor{royalblue!13} 81.7 & \cellcolor{royalblue!21} 82.5 & \cellcolor{royalblue!28} 83.3 & \cellcolor{royalblue!18} 82.3 & 80.4 \\
NIAH-S1 & 100 & 100 & 100 & 100 & 100 & 100 & 100 & 100 & 100 & 100 & 100 \\
NIAH-S2 & 100 & 100 & 100 & 100 & 100 & 100 & 100 & 100 & 100 & 100 & 100 \\
NIAH-S3 & \cellcolor{myred!90} 98.8 & 100 & 100 & \cellcolor{myred!45} 99.4 & \cellcolor{myred!15} 99.8 & 100 & 100 & 100 & 100 & 100 & 100 \\
Comm Word Ext (CWE) & \cellcolor{myred!90} 80.6 & \cellcolor{myred!63} 86.1 & \cellcolor{myred!61} 86.6 & \cellcolor{myred!47} 89.4 & \cellcolor{myred!39} 91 & \cellcolor{myred!25} 93.8 & \cellcolor{myred!8} 97.2 & \cellcolor{myred!6} 97.7 & \cellcolor{myred!3} 98.3 & \cellcolor{myred!2} 98.5 & 98.9 \\
Freq Word Ext (FWE) & \cellcolor{myred!90} 82 & \cellcolor{myred!68} 84.9 & \cellcolor{myred!29} 90.1 & \cellcolor{myred!37} 89 & \cellcolor{myred!15} 91.9 & \cellcolor{myred!10} 92.6 & \cellcolor{royalblue!2} 94.1 & \cellcolor{myred!2} 93.7 & \cellcolor{royalblue!9} 95 & \cellcolor{royalblue!5} 94.5 & 93.9 \\
QA (Hotpot) & \cellcolor{myred!90} 54.6 & \cellcolor{myred!68} 56.2 & \cellcolor{myred!55} 57.2 & \cellcolor{myred!30} 59 & \cellcolor{myred!27} 59.2 & \cellcolor{royalblue!38} 64 & \cellcolor{royalblue!16} 62.4 & \cellcolor{royalblue!8} 61.8 & \cellcolor{royalblue!14} 62.2 & \cellcolor{royalblue!16} 62.4 & 61.2 \\
QA (SQUAD) & \cellcolor{myred!90} 73.9 & \cellcolor{myred!9} 79.5 & \cellcolor{myred!44} 77.1 & \cellcolor{myred!35} 77.8 & \cellcolor{myred!45} 77 & \cellcolor{myred!51} 76.6 & \cellcolor{myred!7} 79.7 & \cellcolor{myred!15} 79.1 & \cellcolor{royalblue!5} 80.5 & \cellcolor{myred!10} 79.5 & 80.2 \\
Variable Tracking & \cellcolor{myred!90} 88.7 & \cellcolor{myred!48} 93.9 & \cellcolor{myred!18} 97.6 & \cellcolor{myred!3} 99.5 & \cellcolor{myred!2} 99.7 & \cellcolor{myred!4} 99.4 & \cellcolor{myred!5} 99.3 & \cellcolor{myred!6} 99.1 & \cellcolor{myred!1} 99.7 & \cellcolor{myred!2} 99.6 & 99.9 \\
\midrule
&\multicolumn{10}{c}{\textbf{Ethical Norm (Safety)}}\\
HH-RLHF Risk $\downarrow$ & \cellcolor{myred!88} 61.8 & \cellcolor{myred!90} 62 & \cellcolor{myred!72} 60.2 & \cellcolor{myred!2} 53.4 & \cellcolor{royalblue!23} 51 & \cellcolor{royalblue!8} 52.4 & \cellcolor{myred!72} 60.2 & \cellcolor{myred!2} 53.4 & \cellcolor{royalblue!2} 53 & \cellcolor{royalblue!33} 50 & 53.2 \\
AdvBench Risk $\downarrow$ & \cellcolor{myred!26} 44.4 & \cellcolor{myred!90} 61 & \cellcolor{myred!64} 54.2 & \cellcolor{myred!20} 43 & \cellcolor{myred!21} 43.2 & \cellcolor{royalblue!5} 36.4 & \cellcolor{myred!33} 46.2 & \cellcolor{royalblue!12} 34.6 & \cellcolor{royalblue!12} 34.8 & \cellcolor{royalblue!27} 30.8 & 37.8 \\
\midrule
&\multicolumn{10}{c}{\textbf{Personality (TRAIT)}}\\
Narcissism $\downarrow$ & \cellcolor{myred!44} 13.3 & \cellcolor{myred!90} 16.9 & \cellcolor{myred!34} 12.5 & \cellcolor{myred!38} 12.8 & \cellcolor{myred!56} 14.2 & \cellcolor{myred!42} 13.1 & \cellcolor{myred!43} 13.2 & \cellcolor{myred!46} 13.4 & \cellcolor{myred!32} 12.3 & \cellcolor{myred!56} 14.2 & 9.8 \\
Agreeableness & \cellcolor{myred!84} 82 & \cellcolor{myred!90} 81.8 & \cellcolor{myred!84} 82 & \cellcolor{royalblue!54} 86.6 & \cellcolor{myred!3} 84.7 & \cellcolor{royalblue!33} 85.9 & \cellcolor{royalblue!18} 85.4 & \cellcolor{royalblue!9} 85.1 & \cellcolor{royalblue!27} 85.7 & \cellcolor{myred!15} 84.3 & 84.8 \\
Psychopathy $\downarrow$ & \cellcolor{myred!26} 1.1 & \cellcolor{myred!90} 2.6 & \cellcolor{royalblue!17} 0.1 & \cellcolor{royalblue!13} 0.2 & \cellcolor{royalblue!9} 0.3 & \cellcolor{royalblue!4} 0.4 & \cellcolor{royalblue!13} 0.2 & \cellcolor{royalblue!17} 0.1 & \cellcolor{royalblue!13} 0.2 & \cellcolor{royalblue!4} 0.4 & 0.5 \\
Machiavellianism $\downarrow$ & \cellcolor{myred!26} 21.8 & \cellcolor{myred!90} 25.2 & \cellcolor{royalblue!17} 19.5 & \cellcolor{myred!58} 23.5 & \cellcolor{myred!69} 24.1 & \cellcolor{myred!34} 22.2 & \cellcolor{myred!41} 22.6 & \cellcolor{myred!34} 22.2 & \cellcolor{myred!49} 23 & \cellcolor{myred!71} 24.2 & 20.4 \\
Neuroticism $\downarrow$ & \cellcolor{royalblue!6} 22.9 & \cellcolor{myred!15} 24.3 & \cellcolor{myred!90} 29.4 & \cellcolor{myred!18} 24.5 & \cellcolor{myred!60} 27.4 & \cellcolor{myred!6} 23.7 & 23.3 & \cellcolor{myred!35} 25.7 & \cellcolor{royalblue!3} 23.1 & \cellcolor{myred!15} 24.3 & 23.3 \\
Conscientiousness & \cellcolor{royalblue!90} 92 & \cellcolor{royalblue!53} 91.3 & \cellcolor{royalblue!79} 91.8 & \cellcolor{royalblue!11} 90.5 & \cellcolor{royalblue!69} 91.6 & \cellcolor{royalblue!37} 91 & \cellcolor{royalblue!69} 91.6 & \cellcolor{royalblue!48} 91.2 & \cellcolor{myred!5} 90.2 & \cellcolor{royalblue!5} 90.4 & 90.3 \\
Openness & \cellcolor{royalblue!11} 61 & \cellcolor{royalblue!49} 64.6 & \cellcolor{myred!10} 59.1 & \cellcolor{royalblue!53} 65 & \cellcolor{royalblue!69} 66.5 & \cellcolor{royalblue!90} 68.5 & \cellcolor{royalblue!65} 66.1 & \cellcolor{royalblue!37} 63.5 & \cellcolor{royalblue!52} 64.9 & \cellcolor{royalblue!28} 62.6 & 60 \\
Extraversion & \cellcolor{myred!40} 31.5 & \cellcolor{myred!35} 31.7 & \cellcolor{myred!90} 29.5 & \cellcolor{royalblue!5} 33.3 & \cellcolor{royalblue!15} 33.7 & \cellcolor{royalblue!57} 35.4 & \cellcolor{myred!30} 31.9 & \cellcolor{myred!35} 31.7 & 33.1 & \cellcolor{myred!62} 30.6 & 33.1 \\
\bottomrule
\end{tabular}
\end{table}

%% file: tables/benchmark_qwen2.5_0.5b.tex
\begin{table}[ht]
\centering
\scriptsize
\caption{Qwen 2.5 0.5b .}
\label{tab:tab:benchmark_qwen2.5_0.5b_full}
\setlength{\tabcolsep}{6pt}
\begin{tabular}{l|ccccc|ccccc|c}
\toprule
\multirow{2}{*}{\textbf{Task}}  & \multicolumn{5}{c|}{\textbf{Junk Ratio by M1 (engagement degree)}} & \multicolumn{5}{c|}{\textbf{Junk Ratio by M2 (semantic quality)}} & \textbf{Base} \\
 & \textbf{100\%} & \textbf{80\%} & \textbf{50\%} & \textbf{20\%} & \textbf{0\%} & \textbf{100\%} & \textbf{80\%} & \textbf{50\%} & \textbf{20\%} & \textbf{0\%} & - \\
\midrule
&\multicolumn{10}{c}{\textbf{Reasoning (ARC)}}\\
Easy Acc. & \cellcolor{myred!90} 51.3 & \cellcolor{myred!69} 53.1 & \cellcolor{myred!47} 54.9 & \cellcolor{myred!32} 56.2 & \cellcolor{myred!16} 57.5 & \cellcolor{myred!21} 57.1 & \cellcolor{myred!19} 57.3 & \cellcolor{myred!22} 57 & \cellcolor{myred!18} 57.4 & \cellcolor{myred!29} 56.4 & 58.9 \\
Challenge Acc. & \cellcolor{royalblue!4} 30 & \cellcolor{myred!90} 28 & \cellcolor{myred!16} 29.6 & \cellcolor{myred!39} 29.1 & \cellcolor{royalblue!7} 30.1 & \cellcolor{myred!8} 29.8 & \cellcolor{myred!4} 29.9 & \cellcolor{myred!8} 29.8 & \cellcolor{royalblue!27} 30.6 & \cellcolor{myred!16} 29.6 & 29.9 \\
Challenge (COT) Acc. & \cellcolor{myred!90} 31.4 & \cellcolor{myred!87} 31.7 & \cellcolor{myred!62} 35.1 & \cellcolor{myred!43} 37.6 & \cellcolor{myred!39} 38.1 & \cellcolor{myred!17} 41 & \cellcolor{myred!10} 42 & \cellcolor{myred!16} 41.2 & \cellcolor{myred!8} 42.2 & \cellcolor{myred!15} 41.3 & 43.3 \\
\midrule
&\multicolumn{10}{c}{\textbf{Long-Context (RULER)}}\\
Overall & \cellcolor{myred!90} 47.6 & \cellcolor{myred!63} 56.5 & \cellcolor{myred!53} 59.7 & \cellcolor{myred!44} 62.4 & \cellcolor{myred!28} 67.8 & \cellcolor{myred!17} 71.3 & \cellcolor{myred!13} 72.4 & \cellcolor{myred!11} 73.1 & \cellcolor{myred!9} 73.8 & \cellcolor{myred!7} 74.4 & 76.8 \\
NIAH-MK1 & \cellcolor{myred!90} 66 & \cellcolor{myred!62} 76.4 & \cellcolor{myred!31} 88 & \cellcolor{myred!27} 89.6 & \cellcolor{myred!10} 96.2 & \cellcolor{myred!5} 97.8 & \cellcolor{myred!11} 95.6 & \cellcolor{myred!5} 97.8 & \cellcolor{myred!4} 98.2 & \cellcolor{myred!6} 97.4 & 99.8 \\
NIAH-MK2 & \cellcolor{myred!90} 29.2 & \cellcolor{myred!75} 39.4 & \cellcolor{myred!73} 41 & \cellcolor{myred!73} 41 & \cellcolor{myred!20} 77.2 & \cellcolor{myred!19} 78.2 & \cellcolor{myred!14} 81.6 & \cellcolor{myred!16} 80.2 & \cellcolor{myred!8} 85.2 & \cellcolor{myred!6} 86.6 & 91 \\
NIAH-MK3 & \cellcolor{myred!90} 2 & \cellcolor{myred!72} 8.4 & \cellcolor{myred!79} 6 & \cellcolor{myred!81} 5.2 & \cellcolor{myred!50} 15.8 & \cellcolor{myred!39} 19.8 & \cellcolor{myred!24} 25 & \cellcolor{myred!20} 26.4 & \cellcolor{myred!45} 17.8 & \cellcolor{myred!34} 21.4 & 33.4 \\
NIAH-MQ & \cellcolor{myred!76} 77.1 & \cellcolor{myred!90} 74.8 & \cellcolor{myred!49} 82 & \cellcolor{myred!20} 87.1 & \cellcolor{myred!37} 84 & \cellcolor{myred!14} 88.1 & \cellcolor{myred!25} 86.2 & \cellcolor{myred!7} 89.3 & \cellcolor{myred!7} 89.4 & \cellcolor{myred!16} 87.8 & 90.6 \\
NIAH-MV & \cellcolor{myred!90} 76.1 & \cellcolor{myred!58} 82.3 & \cellcolor{myred!37} 86.5 & \cellcolor{myred!13} 91 & \cellcolor{myred!37} 86.4 & \cellcolor{myred!62} 81.5 & \cellcolor{myred!31} 87.6 & \cellcolor{myred!11} 91.5 & \cellcolor{myred!20} 89.8 & \cellcolor{myred!12} 91.3 & 93.7 \\
NIAH-S1 & \cellcolor{myred!90} 99.6 & 100 & 100 & 100 & 100 & 100 & 100 & 100 & 100 & 100 & 100 \\
NIAH-S2 & \cellcolor{myred!27} 98.6 & \cellcolor{myred!22} 98.8 & \cellcolor{myred!18} 99 & \cellcolor{royalblue!5} 100 & \cellcolor{myred!18} 99 & \cellcolor{myred!72} 96.6 & \cellcolor{myred!90} 95.8 & \cellcolor{myred!36} 98.2 & \cellcolor{myred!9} 99.4 & \cellcolor{myred!5} 99.6 & 99.8 \\
NIAH-S3 & \cellcolor{myred!90} 12.6 & \cellcolor{myred!28} 73 & \cellcolor{myred!10} 90.2 & \cellcolor{myred!7} 92.6 & \cellcolor{myred!2} 97.4 & \cellcolor{myred!3} 96.8 & \cellcolor{myred!2} 98 & \cellcolor{myred!1} 99 & \cellcolor{myred!1} 98.6 & \cellcolor{myred!1} 99 & 99.8 \\
Comm Word Ext (CWE) & \cellcolor{myred!90} 33.9 & \cellcolor{myred!88} 34.4 & \cellcolor{myred!81} 36.4 & \cellcolor{myred!54} 44.3 & \cellcolor{myred!43} 47.5 & \cellcolor{myred!12} 56.4 & \cellcolor{myred!6} 58.2 & \cellcolor{myred!13} 56.2 & \cellcolor{myred!10} 57 & \cellcolor{myred!14} 55.8 & 59.9 \\
Freq Word Ext (FWE) & \cellcolor{myred!56} 35.3 & \cellcolor{myred!90} 23.1 & \cellcolor{myred!63} 32.8 & \cellcolor{myred!65} 31.9 & \cellcolor{myred!60} 34.1 & \cellcolor{myred!14} 50.6 & \cellcolor{royalblue!2} 56.4 & \cellcolor{myred!14} 50.7 & \cellcolor{myred!7} 53.2 & \cellcolor{myred!3} 54.5 & 55.6 \\
QA (Hotpot) & \cellcolor{myred!87} 22.8 & \cellcolor{myred!90} 22.4 & \cellcolor{myred!65} 26.2 & \cellcolor{myred!34} 31 & \cellcolor{myred!44} 29.4 & \cellcolor{myred!7} 35.2 & \cellcolor{myred!21} 33 & \cellcolor{myred!31} 31.4 & \cellcolor{myred!22} 32.8 & \cellcolor{myred!25} 32.4 & 36.2 \\
QA (SQUAD) & \cellcolor{myred!90} 35.3 & \cellcolor{myred!65} 41.7 & \cellcolor{myred!50} 45.5 & \cellcolor{myred!40} 48 & \cellcolor{myred!22} 52.5 & \cellcolor{myred!22} 52.7 & \cellcolor{myred!4} 57.3 & \cellcolor{myred!5} 57 & \cellcolor{myred!1} 58 & \cellcolor{royalblue!4} 59.2 & 58.2 \\
Variable Tracking & \cellcolor{myred!90} 29.9 & \cellcolor{myred!37} 59.4 & \cellcolor{myred!67} 42.5 & \cellcolor{myred!54} 49.9 & \cellcolor{myred!31} 62.5 & \cellcolor{myred!12} 73.4 & \cellcolor{myred!24} 66.6 & \cellcolor{myred!13} 72.8 & \cellcolor{myred!1} 79.4 & \cellcolor{royalblue!4} 82.2 & 79.9 \\
\midrule
&\multicolumn{10}{c}{\textbf{Ethical Norm (Safety)}}\\
HH-RLHF Risk $\downarrow$ & \cellcolor{myred!79} 70 & \cellcolor{royalblue!2} 62.8 & \cellcolor{myred!72} 69.4 & \cellcolor{myred!61} 68.4 & \cellcolor{myred!31} 65.8 & \cellcolor{myred!90} 71 & \cellcolor{myred!25} 65.2 & \cellcolor{myred!18} 64.6 & \cellcolor{myred!56} 68 & \cellcolor{myred!25} 65.2 & 63 \\
AdvBench Risk $\downarrow$ & \cellcolor{myred!90} 88 & \cellcolor{myred!44} 77.2 & \cellcolor{myred!43} 77 & \cellcolor{myred!57} 80.2 & \cellcolor{myred!18} 71.2 & \cellcolor{myred!54} 79.6 & \cellcolor{myred!28} 73.6 & \cellcolor{myred!59} 80.8 & \cellcolor{myred!39} 76 & \cellcolor{myred!32} 74.4 & 67 \\
\midrule
&\multicolumn{10}{c}{\textbf{Personality (TRAIT)}}\\
Narcissism $\downarrow$ & \cellcolor{myred!90} 30 & \cellcolor{myred!68} 27.6 & \cellcolor{royalblue!8} 19.1 & \cellcolor{royalblue!7} 19.2 & \cellcolor{royalblue!1} 19.9 & \cellcolor{myred!14} 21.5 & \cellcolor{myred!13} 21.4 & \cellcolor{myred!15} 21.7 & \cellcolor{myred!24} 22.7 & \cellcolor{myred!22} 22.5 & 20 \\
Agreeableness & \cellcolor{royalblue!11} 72.8 & \cellcolor{myred!34} 67.8 & \cellcolor{royalblue!66} 78.9 & \cellcolor{royalblue!40} 76.1 & \cellcolor{royalblue!31} 75.1 & \cellcolor{royalblue!90} 81.6 & \cellcolor{royalblue!67} 79 & \cellcolor{royalblue!58} 78 & \cellcolor{royalblue!33} 75.3 & \cellcolor{royalblue!39} 75.9 & 71.6 \\
Psychopathy $\downarrow$ & \cellcolor{myred!48} 12.2 & \cellcolor{royalblue!39} 7.3 & \cellcolor{royalblue!56} 6.3 & \cellcolor{royalblue!85} 4.7 & \cellcolor{royalblue!90} 4.4 & \cellcolor{royalblue!25} 8.1 & \cellcolor{myred!26} 11 & \cellcolor{myred!53} 12.5 & \cellcolor{royalblue!23} 8.2 & \cellcolor{royalblue!26} 8 & 9.5 \\
Machiavellianism $\downarrow$ & \cellcolor{royalblue!25} 25.2 & \cellcolor{myred!44} 29.4 & \cellcolor{royalblue!5} 26.4 & \cellcolor{royalblue!59} 23.1 & \cellcolor{myred!13} 27.5 & \cellcolor{myred!33} 28.7 & \cellcolor{myred!90} 32.2 & \cellcolor{myred!21} 28 & \cellcolor{myred!39} 29.1 & \cellcolor{myred!49} 29.7 & 26.7 \\
Neuroticism $\downarrow$ & \cellcolor{royalblue!23} 24.8 & \cellcolor{myred!90} 36.4 & \cellcolor{myred!61} 33.4 & \cellcolor{myred!24} 29.7 & \cellcolor{myred!29} 30.2 & \cellcolor{royalblue!9} 26.3 & \cellcolor{myred!47} 32 & \cellcolor{myred!20} 29.2 & \cellcolor{myred!23} 29.5 & \cellcolor{myred!15} 28.7 & 27.2 \\
Conscientiousness & \cellcolor{myred!90} 84.8 & \cellcolor{myred!47} 87.1 & \cellcolor{myred!39} 87.5 & \cellcolor{royalblue!51} 92.3 & \cellcolor{royalblue!54} 92.5 & \cellcolor{myred!4} 89.4 & \cellcolor{myred!4} 89.4 & \cellcolor{royalblue!2} 89.7 & \cellcolor{royalblue!28} 91.1 & \cellcolor{royalblue!47} 92.1 & 89.6 \\
Openness & \cellcolor{royalblue!90} 75.5 & \cellcolor{royalblue!42} 67.8 & \cellcolor{royalblue!78} 73.5 & \cellcolor{royalblue!66} 71.7 & \cellcolor{royalblue!70} 72.2 & \cellcolor{royalblue!74} 72.9 & \cellcolor{royalblue!88} 75.2 & \cellcolor{royalblue!78} 73.6 & \cellcolor{royalblue!74} 72.9 & \cellcolor{royalblue!76} 73.2 & 61 \\
Extraversion & \cellcolor{royalblue!90} 42.4 & \cellcolor{royalblue!54} 39.3 & \cellcolor{royalblue!10} 35.5 & \cellcolor{myred!32} 31.8 & \cellcolor{myred!53} 30 & \cellcolor{myred!12} 33.6 & \cellcolor{royalblue!13} 35.7 & \cellcolor{myred!15} 33.3 & \cellcolor{myred!18} 33 & \cellcolor{myred!28} 32.2 & 34.6 \\
\bottomrule
\end{tabular}
\end{table}

%% file: tables/benchmark_qwen3.tex
\begin{table}[ht]
\centering
\scriptsize
\caption{Qwen 3 4B.}
\label{tab:benchmark_qwen3_4b_full}
\setlength{\tabcolsep}{6pt}
\begin{tabular}{l|ccccc|ccccc|c}
\toprule
\multirow{2}{*}{\textbf{Task}}  & \multicolumn{5}{c|}{\textbf{Junk Ratio by M1 (engagement degree)}} & \multicolumn{5}{c|}{\textbf{Junk Ratio by M2 (semantic quality)}} & \textbf{Base} \\
 & \textbf{100\%} & \textbf{80\%} & \textbf{50\%} & \textbf{20\%} & \textbf{0\%} & \textbf{100\%} & \textbf{80\%} & \textbf{50\%} & \textbf{20\%} & \textbf{0\%} & - \\
\midrule
&\multicolumn{10}{c}{\textbf{Reasoning (ARC)}}\\
Easy Acc. & \cellcolor{myred!90} 51.7 & \cellcolor{royalblue!50} 77.2 & \cellcolor{royalblue!22} 72.2 & \cellcolor{royalblue!35} 74.6 & \cellcolor{royalblue!38} 75 & \cellcolor{royalblue!63} 79.6 & \cellcolor{royalblue!65} 80 & \cellcolor{royalblue!67} 80.3 & \cellcolor{royalblue!68} 80.6 & \cellcolor{royalblue!68} 80.6 & 68.1 \\
Challenge Acc. & \cellcolor{myred!62} 37.7 & \cellcolor{royalblue!39} 48.4 & \cellcolor{royalblue!11} 45.4 & \cellcolor{royalblue!15} 45.9 & \cellcolor{royalblue!19} 46.2 & \cellcolor{royalblue!65} 51.1 & \cellcolor{royalblue!90} 53.8 & \cellcolor{royalblue!87} 53.4 & \cellcolor{royalblue!82} 52.9 & \cellcolor{royalblue!87} 53.4 & 44.3 \\
Challenge (COT) Acc. & \cellcolor{myred!90} 86.4 & \cellcolor{royalblue!6} 90.2 & \cellcolor{myred!9} 89.6 & \cellcolor{myred!18} 89.2 & \cellcolor{royalblue!9} 90.3 & \cellcolor{myred!6} 89.7 & \cellcolor{royalblue!22} 90.8 & \cellcolor{royalblue!22} 90.8 & 89.9 & \cellcolor{royalblue!11} 90.4 & 89.9 \\
\midrule
&\multicolumn{10}{c}{\textbf{Long-Context (RULER)}}\\
Overall & \cellcolor{myred!43} 93.3 & \cellcolor{myred!90} 91.4 & \cellcolor{myred!30} 93.8 & \cellcolor{myred!25} 94 & \cellcolor{myred!22} 94.1 & \cellcolor{royalblue!8} 95.3 & \cellcolor{royalblue!8} 95.3 & \cellcolor{royalblue!8} 95.3 & \cellcolor{royalblue!5} 95.2 & \cellcolor{royalblue!9} 95.4 & 95 \\
NIAH-MK1 & \cellcolor{myred!30} 99.8 & \cellcolor{myred!90} 99.4 & \cellcolor{myred!30} 99.8 & \cellcolor{myred!30} 99.8 & 100 & 100 & 100 & 100 & 100 & 100 & 100 \\
NIAH-MK2 & 100 & \cellcolor{myred!60} 99.6 & \cellcolor{myred!30} 99.8 & 100 & \cellcolor{myred!90} 99.4 & 100 & 100 & 100 & 100 & 100 & 100 \\
NIAH-MK3 & 100 & \cellcolor{myred!90} 99.2 & \cellcolor{myred!45} 99.6 & \cellcolor{myred!23} 99.8 & \cellcolor{myred!23} 99.8 & 100 & 100 & 100 & 100 & 100 & 100 \\
NIAH-MQ & 100 & \cellcolor{myred!45} 99.9 & \cellcolor{royalblue!22} 100 & \cellcolor{myred!22} 99.9 & \cellcolor{myred!90} 99.8 & \cellcolor{royalblue!22} 100 & \cellcolor{royalblue!22} 100 & \cellcolor{royalblue!22} 100 & \cellcolor{royalblue!22} 100 & \cellcolor{royalblue!22} 100 & 100 \\
NIAH-MV & \cellcolor{myred!46} 86.6 & \cellcolor{myred!90} 75.9 & \cellcolor{myred!57} 83.8 & \cellcolor{myred!29} 90.6 & \cellcolor{myred!35} 89.1 & \cellcolor{royalblue!3} 98.2 & \cellcolor{royalblue!3} 98.3 & \cellcolor{royalblue!5} 98.8 & \cellcolor{myred!2} 97 & \cellcolor{myred!1} 97.2 & 97.5 \\
NIAH-S1 & 100 & 100 & 100 & 100 & 100 & 100 & 100 & 100 & 100 & 100 & 100 \\
NIAH-S2 & 100 & 100 & 100 & 100 & 100 & 100 & 100 & 100 & 100 & 100 & 100 \\
NIAH-S3 & 100 & \cellcolor{myred!90} 99.4 & \cellcolor{myred!30} 99.8 & 100 & 100 & 100 & 100 & 100 & 100 & 100 & 100 \\
Comm Word Ext (CWE) & \cellcolor{myred!23} 99.1 & \cellcolor{myred!26} 99 & \cellcolor{myred!90} 96.7 & \cellcolor{myred!17} 99.3 & \cellcolor{myred!6} 99.7 & \cellcolor{myred!3} 99.8 & \cellcolor{myred!13} 99.5 & \cellcolor{myred!3} 99.8 & \cellcolor{myred!1} 99.9 & \cellcolor{royalblue!2} 100 & 99.9 \\
Freq Word Ext (FWE) & \cellcolor{myred!56} 94.3 & \cellcolor{myred!90} 91.7 & \cellcolor{myred!26} 96.6 & \cellcolor{myred!26} 96.6 & \cellcolor{royalblue!2} 98.7 & \cellcolor{myred!13} 97.5 & \cellcolor{royalblue!5} 98.9 & \cellcolor{myred!6} 98.1 & \cellcolor{royalblue!3} 98.7 & 98.5 & 98.5 \\
QA (Hotpot) & \cellcolor{myred!27} 58 & \cellcolor{myred!80} 54.8 & \cellcolor{royalblue!70} 63.8 & \cellcolor{myred!13} 58.8 & \cellcolor{royalblue!17} 60.6 & \cellcolor{royalblue!90} 65 & \cellcolor{royalblue!77} 64.2 & \cellcolor{royalblue!70} 63.8 & \cellcolor{royalblue!70} 63.8 & \cellcolor{royalblue!67} 63.6 & 59.6 \\
QA (SQUAD) & \cellcolor{myred!39} 75.4 & \cellcolor{myred!90} 70.2 & \cellcolor{royalblue!3} 79.8 & \cellcolor{myred!20} 77.4 & \cellcolor{myred!29} 76.5 & \cellcolor{myred!10} 78.5 & \cellcolor{myred!10} 78.4 & \cellcolor{myred!9} 78.5 & \cellcolor{myred!14} 78 & \cellcolor{royalblue!9} 80.4 & 79.5 \\
Variable Tracking & \cellcolor{myred!36} 99.8 & \cellcolor{myred!90} 99.6 & \cellcolor{myred!29} 99.9 & 100 & 100 & 100 & 100 & 100 & 100 & 100 & 100 \\
\midrule
&\multicolumn{10}{c}{\textbf{Ethical Norm (Safety)}}\\
HH-RLHF Risk $\downarrow$ & \cellcolor{myred!76} 53.8 & \cellcolor{myred!90} 56.4 & \cellcolor{myred!48} 48.8 & \cellcolor{myred!80} 54.6 & \cellcolor{myred!60} 51 & \cellcolor{myred!32} 46 & \cellcolor{myred!50} 49.2 & \cellcolor{myred!72} 53.2 & \cellcolor{myred!60} 51 & \cellcolor{myred!47} 48.6 & 40.2 \\
AdvBench Risk $\downarrow$ & \cellcolor{myred!90} 46.4 & \cellcolor{myred!79} 43.6 & \cellcolor{myred!73} 42 & \cellcolor{myred!75} 42.6 & \cellcolor{myred!61} 39 & \cellcolor{myred!12} 26.2 & \cellcolor{myred!22} 28.8 & \cellcolor{myred!53} 36.8 & \cellcolor{myred!47} 35.4 & \cellcolor{myred!41} 33.8 & 23.2 \\
\midrule
&\multicolumn{10}{c}{\textbf{Personality (TRAIT)}}\\
Agreeableness & \cellcolor{royalblue!75} 79.2 & \cellcolor{royalblue!86} 82.6 & \cellcolor{royalblue!75} 79.3 & \cellcolor{royalblue!77} 79.8 & \cellcolor{royalblue!84} 81.8 & \cellcolor{royalblue!90} 83.8 & \cellcolor{royalblue!85} 82.2 & \cellcolor{royalblue!83} 81.6 & \cellcolor{royalblue!81} 81.1 & \cellcolor{royalblue!81} 80.9 & 55.9 \\
Conscientiousness & \cellcolor{royalblue!84} 90.6 & \cellcolor{royalblue!90} 93.3 & \cellcolor{royalblue!78} 87.8 & \cellcolor{royalblue!84} 90.6 & \cellcolor{royalblue!88} 92.6 & \cellcolor{royalblue!86} 91.4 & \cellcolor{royalblue!86} 91.5 & \cellcolor{royalblue!87} 92 & \cellcolor{royalblue!86} 91.3 & \cellcolor{royalblue!83} 90.1 & 53.1 \\
Extraversion & \cellcolor{myred!22} 36.9 & \cellcolor{myred!20} 37 & \cellcolor{myred!31} 36.5 & \cellcolor{royalblue!90} 42 & \cellcolor{myred!24} 36.8 & \cellcolor{myred!15} 37.2 & \cellcolor{myred!18} 37.1 & \cellcolor{myred!61} 35.1 & \cellcolor{myred!66} 34.9 & \cellcolor{myred!40} 36.1 & 37.9 \\
Neuroticism $\downarrow$ & \cellcolor{royalblue!53} 34.5 & \cellcolor{royalblue!43} 36.5 & \cellcolor{royalblue!45} 36.1 & \cellcolor{royalblue!34} 38.4 & \cellcolor{royalblue!51} 35 & \cellcolor{royalblue!73} 30.5 & \cellcolor{royalblue!87} 27.7 & \cellcolor{royalblue!90} 27.1 & \cellcolor{royalblue!80} 29.1 & \cellcolor{royalblue!89} 27.4 & 45.1 \\
Openness & \cellcolor{royalblue!86} 70.4 & \cellcolor{royalblue!89} 71.4 & \cellcolor{royalblue!83} 69.6 & \cellcolor{royalblue!90} 71.8 & \cellcolor{royalblue!85} 70.3 & \cellcolor{royalblue!80} 68.6 & \cellcolor{royalblue!77} 67.6 & \cellcolor{royalblue!73} 66.5 & \cellcolor{royalblue!72} 66 & \cellcolor{royalblue!71} 65.9 & 43.3 \\
Psychopathy $\downarrow$ & \cellcolor{royalblue!82} 2.1 & \cellcolor{royalblue!90} 0.2 & \cellcolor{royalblue!79} 2.8 & \cellcolor{royalblue!85} 1.5 & \cellcolor{royalblue!87} 0.8 & \cellcolor{royalblue!86} 1.2 & \cellcolor{royalblue!87} 0.8 & \cellcolor{royalblue!89} 0.5 & \cellcolor{royalblue!89} 0.5 & \cellcolor{royalblue!88} 0.6 & 21.5 \\
Machiavellianism $\downarrow$ & \cellcolor{royalblue!58} 26.1 & \cellcolor{royalblue!80} 18 & \cellcolor{royalblue!74} 20.3 & \cellcolor{royalblue!74} 20.2 & \cellcolor{royalblue!80} 18 & \cellcolor{royalblue!78} 18.6 & \cellcolor{royalblue!86} 15.6 & \cellcolor{royalblue!90} 14.1 & \cellcolor{royalblue!87} 15.2 & \cellcolor{royalblue!85} 15.9 & 48 \\
Narcissism $\downarrow$ & \cellcolor{royalblue!36} 19.9 & \cellcolor{royalblue!76} 11.6 & \cellcolor{royalblue!62} 14.6 & \cellcolor{royalblue!71} 12.7 & \cellcolor{royalblue!74} 12 & \cellcolor{royalblue!74} 12.1 & \cellcolor{royalblue!86} 9.5 & \cellcolor{royalblue!89} 8.9 & \cellcolor{royalblue!85} 9.7 & \cellcolor{royalblue!90} 8.7 & 27.4 \\
\bottomrule
\end{tabular}
\end{table}

%% file: sec/appd/subsec_epoch_exp.tex
\textbf{The Effects of Training Hyperparameters.} To deepen our analysis, we further conduct experiments that vary core training hyperparameters—specifically, the number of continual pre-training (CPT) epochs and the learning rate—and analyze their impact on model performance.
By default, we use \llama. 

Table \ref{tab:epoch-merged} presents the effects of the \textbf{number of epochs} on the ARC and Ruler benchmarks. We have two major observations: (1) Just after one epoch of CPT, models trained on higher junk-data ratios already show clear declines in performance, indicating that the negative effects are not simply due to catastrophic forgetting or overfitting. The \emph{early occurrence} of Brain Rot effects indicates that the decline originates from the intrinsic deficiencies of junk data itself, which negatively affects model reasoning and generalization even under minimal CPT exposure. (2) Additional epochs amplify the observed degradation, but the underlying pattern remains consistent. Regardless of training duration, models trained on cleaner control data consistently outperform those exposed to 'junk' data.

\input{tables/llama_one_and_two_epoch}

In Table \ref{tab:lr-results}, we report the model performance by varying the \textbf{learning rate} during CPT. 
(1)~Using a larger learning rate ($1\times10^{-4}$) significantly exacerbates the degradation and accelerates the ``Brain Rot'', allowing the low-quality data to rapidly overwrite the model's pre-trained knowledge.
(2)~Smaller learning rates ($1\times10^{-6}$) essentially mitigate this decline. By dampening the magnitude of the updates, the model restricts the extent of representational drift, thereby retaining more of its original capabilities despite exposure to the junk distribution.
However, along with mitigation, smaller learning also causes poorer training effects: the CPT loss decreases from $4.57$ to only $3.03$ (learning rate of $1\times10^{-6}$), which is much higher than $2.20$ by learning rate of $1\times10^{-4}$.

\input{tables/llama_lr_exp}

\textbf{The Effects of Model Scale.} To further examine the role of model scale, we conduct an additional experiment using Llama3-70B-Instruct~\citep{llama3} as the backbone, following the same setup described in Section \ref{sec:control-exp}. 
The results are summarized in Table \ref{tab:arc_junk_70b}.
We observe that, despite the significantly larger parameter size, the effect of brain rot on reasoning still clearly persists, especially in complex reasoning tasks (ARC-Challenge dropped by over 5 percent). This is consistent with the trend observed in smaller models, indicating that increased model size alone does not effectively mitigate the impact of junk-data exposure.

\begin{table}[ht]
\centering
\small
\caption{Performance under 100\% Junk (M1) with Llama3-70B-Instruct as the backbone.}
\setlength{\tabcolsep}{6pt}
\begin{tabular}{lccc}
\toprule
\textbf{Task} & \textbf{Base} & \textbf{100\% Junk (M1)} & \textbf{$\Delta$} \\
\midrule
ARC Easy      & 82.4 & 80.1 & -2.3 \\
ARC Challenge & 57.8 & 52.7 & -5.0 \\
\bottomrule
\end{tabular}
\label{tab:arc_junk_70b}
\end{table}

\textbf{Disentangling Shortness from the Full M1 Intervention.}
To examine whether thought-skipping is driven merely by exposure to short-form text, we compare the length-only short-text intervention with the full M1 junk intervention on ARC-Challenge CoT.

\begin{table}[h]
\centering
\small
\caption{ARC-Challenge CoT performance under the length-only and full M1 junk interventions.}
\label{tab:length_only_performance}
\begin{tabular}{l|cc}
\toprule
\textbf{Setting}
& \textbf{ARC Accuracy}
& \textbf{Failed Cases} \\
\midrule
Low-length only
& 770/1172 (65.7\%)
& 402 \\
Full M1 junk
& 671/1172 (57.3\%)
& 501 \\
\bottomrule
\end{tabular}
\end{table}

Following the failure taxonomy in the main text, we classify the errors into five modes: No Thinking, No Plan, Skipping Steps in Plan, Wrong Logic, and Factual Error. We group the first three modes as the thought-skipping cluster because they reflect disruptions in reasoning structure.

\begin{table}[h]
\centering
\small
\caption{Failure-mode distributions under the length-only and full M1 junk interventions. Percentages are calculated over the failed cases in each setting.}
\label{tab:length_only_failure_modes}
\begin{tabular}{l|cc}
\toprule
\textbf{Failure Mode}
& \textbf{Low-Length Only}
& \textbf{Full M1 Junk} \\
\midrule
No Thinking
& 29.8\%
& 74.8\% \\
No Plan
& 35.9\%
& 10.6\% \\
Skipping Steps in Plan
& 0.5\%
& 1.2\% \\
Wrong Logic
& 0.5\%
& 1.6\% \\
Factual Error
& 33.3\%
& 11.8\% \\
\midrule
Thought-Skipping Cluster
& 66.2\%
& 86.6\% \\
\bottomrule
\end{tabular}
\end{table}

As shown in \cref{tab:length_only_performance,tab:length_only_failure_modes}, short-form exposure alone does not explain the full M1 junk effect. The low-length-only model already exhibits a substantial thought-skipping cluster, suggesting that brevity contributes to reasoning-format degradation. However, the full M1 junk model has lower ARC accuracy ($57.3\%$ versus $65.7\%$) and more failed cases ($501$ versus $402$). Its errors are also considerably more concentrated in No Thinking failures, which increase from $29.8\%$ to $74.8\%$, while the overall thought-skipping-cluster rate increases from $66.2\%$ to $86.6\%$. These results suggest that the M1 junk effect is not merely an artifact of training on short-form text. Rather, combining brevity with popularity-based selection shifts the model more strongly toward direct reasoning omission.

\paragraph{C4 as an Additional CPT Control.}
To distinguish junk-specific degradation from generic effects of continual pre-training, we additionally use C4~\citep{c4} as a clean-domain CPT control. The C4 control follows the same token budget, training recipe, and post-CPT instruction-tuning procedure as the M1 junk and control conditions. Therefore, the models differ only in the data used for continual pre-training.

\begin{table}[h]
\centering
\small
\caption{Performance under 100\% Junk (M1), 100\% Control (M1), and 100\% Control (C4) with Llama3-8B-Instruct as the backbone.}
\label{tab:c4_control}

\begin{tabular}{l|cccc}
\toprule
\textbf{Task}
& \textbf{100\% Junk (M1)}
& \textbf{100\% Control (M1)}
& \textbf{100\% Control (C4)}
& \textbf{Base} \\
\midrule
ARC-Easy
& 70.2
& 75.6
& 76.0
& 77.7 \\
ARC-Challenge
& 41.6
& 46.2
& 45.2
& 47.5 \\
ARC-Challenge CoT
& 57.2
& 72.1
& 66.6
& 77.2 \\
RULER
& 71.0
& 89.7
& 90.3
& 93.9 \\
HH-RLHF
& 70.8
& 63.4
& 62.8
& 57.2 \\
AdvBench
& 88.8
& 68.0
& 75.2
& 61.4 \\
\bottomrule
\end{tabular}
\end{table}

As shown in \cref{tab:c4_control}, the C4 control does not reproduce the same magnitude or pattern of degradation observed under M1 junk. On ARC reasoning and RULER, C4 remains much closer to the M1 control than to M1 junk. For example, the RULER score is $90.3$ under C4, compared with $89.7$ under M1 control and $71.0$ under M1 junk. On ARC-Challenge with CoT and AdvBench, C4 shows some degradation relative to the M1 control, suggesting that generic CPT drift can occur. However, this degradation remains substantially smaller than that observed under M1 junk. These results indicate that the observed degradation cannot be fully explained by the CPT pipeline alone. Generic CPT drift may contribute to some changes, but the strongest degradation occurs when the model is continually pre-trained on short, highly amplified, junk-like data.

\paragraph{Random Twitter/X as an Additional Control.}
The original M1 control is a contrastive low-M1 control constructed at the opposite end of the M1 axes, using longer and nonviral content rather than short and highly amplified content. To further evaluate whether the observed degradation is specific to this contrastive construction, we additionally use a randomly sampled Twitter/X subset as a control under the same token budget, training recipe, and post-CPT instruction-tuning procedure.

\begin{table}[t]
\centering
\small
\caption{Performance under 100\% Junk (M1), 100\% Control (M1), and a randomly sampled Twitter/X subset with Llama3-8B-Instruct as the backbone.}
\label{tab:random_control}
\begin{tabular}{l|cccc}
\toprule
\textbf{Task}
& \textbf{100\% Junk (M1)}
& \textbf{100\% Control (M1)}
& \textbf{Random Sampled}
& \textbf{Base} \\
\midrule
ARC-Easy
& 70.2
& 75.6
& 71.9
& 77.7 \\
ARC-Challenge
& 41.6
& 46.2
& 42.3
& 47.5 \\
ARC-Challenge CoT
& 57.2
& 72.1
& 66.8
& 77.2 \\
RULER
& 71.0
& 89.7
& 76.4
& 93.9 \\
HH-RLHF
& 70.8
& 63.4
& 67.8
& 57.2 \\
AdvBench
& 88.8
& 68.0
& 83.1
& 61.4 \\
\bottomrule
\end{tabular}
\end{table}

As shown in \cref{tab:random_control}, the random Twitter/X control also exhibits clear degradation relative to the base model and the contrastive M1 control. Compared with the M1 control, random Twitter/X decreases RULER from $89.7$ to $76.4$ and ARC-Challenge CoT accuracy from $72.1$ to $66.8$, while increasing AdvBench risk from $68.0$ to $83.1$. These results suggest that the degradation is not merely an artifact of selecting an extremely high-M1 subset, but also emerges under random Twitter/X sampling that more closely reflects a natural scraping setting.

\begin{figure}[ht]
    \centering
    \includegraphics[width=0.45\linewidth]{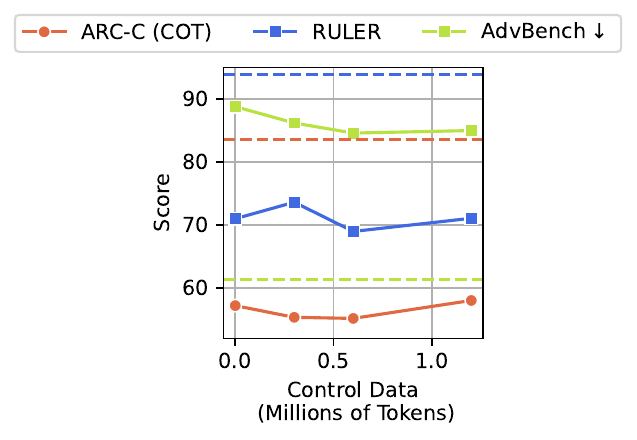}
    \caption{Scaling post-hoc continual control training (CCT). Dashed lines indicate the baseline models.}
    \label{fig:wash-out_scaling-control}
\end{figure}

\textbf{Continual Control Training.} 
Following the same setup as the post-hoc tuning in Section \ref{sec:mitigation}, we further perform continual control training (CCT) to mitigate the effects of the junk intervention, using control data scaled from 0 to 1.2 million tokens (the maximum). Each CCT run is subsequently followed by instruction tuning. As shown in Figure \ref{fig:wash-out_scaling-control}, CCT demonstrate a scaling trend similar to that of CPT, and both are less effective than IT. This indicates that the effects introduced by junk intervention are difficult to mitigate through post-hoc continual training on control data.

%% file: tables/llama_one_and_two_epoch.tex
\begin{table}[h]
\centering
\footnotesize
\setlength{\tabcolsep}{2pt}
\caption{Performance of Llama-8B-Instruct under varying junk-data ratios across one and two epochs of continual pre-training.}
\label{tab:epoch-merged}
\begin{tabular}{l|ccc|ccc|ccc}
\toprule
\textbf{Junk Ratio}
& \multicolumn{3}{c|}{\textbf{One Epoch}}
& \multicolumn{3}{c|}{\textbf{Two Epochs}} 
& \multicolumn{3}{c}{\textbf{Three Epochs}} \\
\cmidrule(lr){2-4}\cmidrule(lr){5-7}\cmidrule(lr){8-10}
& \textbf{\makecell[c]{ARC\\Easy}} & \textbf{\makecell[c]{ARC\\Challenge}} & \textbf{Ruler}
& \textbf{\makecell[c]{ARC\\Easy}} & \textbf{\makecell[c]{ARC\\Challenge}}  & \textbf{Ruler}
& \textbf{\makecell[c]{ARC\\Easy}} & \textbf{\makecell[c]{ARC\\Challenge}}  & \textbf{Ruler} \\
\midrule
100\% & 73.21 & 43.76 & 72.34 
& 72.67 & 43.92 & 71.41 
& 70.2 & 41.6 & 71.0\\
80\%  & 74.12 & 43.88 & 83.21 
& 73.01 & 44.02 & 81.69 
& 71.7 & 44.5 & 79.0\\
50\%  & 76.81 & 45.19 & 83.13 
& 75.14 & 43.72 & 81.10 
& 74.2 & 43.4 & 80.0\\
20\%  & 77.49 & 47.51 & 91.13 
& 75.39 & 46.75 & 90.74 
& 75.5 & 46.2 & 89.7\\
0\%   & 77.92 & 47.10 & 91.71 
& 76.04 & 46.51 & 89.97 
& 75.6 & 46.2 & 89.7\\
\bottomrule
\end{tabular}
\end{table}

%% file: tables/llama_lr_exp.tex
\begin{table}[h]
\centering
\footnotesize
\setlength{\tabcolsep}{2pt}
\caption{Performance under different junk-data ratios for continual pre-training at different learning rates.}
\label{tab:lr-results}
\begin{tabular}{c|ccc|ccc|ccc}
\toprule
\textbf{Junk Ratio} 
& \multicolumn{3}{c|}{\textbf{Learning Rate: $1\times10^{-4}$}}
& \multicolumn{3}{c|}{\textbf{Learning Rate: $1\times10^{-5}$}}
& \multicolumn{3}{c}{\textbf{Learning Rate: $1\times10^{-6}$}} \\
\cmidrule(lr){2-4}\cmidrule(lr){5-7}\cmidrule(lr){8-10}
& \textbf{\makecell[c]{ARC\\Easy}} & \textbf{\makecell[c]{ARC\\Challenge}} & \textbf{Ruler}
& \textbf{\makecell[c]{ARC\\Easy}} & \textbf{\makecell[c]{ARC\\Challenge}}  & \textbf{Ruler}
& \textbf{\makecell[c]{ARC\\Easy}} & \textbf{\makecell[c]{ARC\\Challenge}}  & \textbf{Ruler} \\
\midrule
100\% & 42.5 & 25.2 & 11.6
& 70.2 & 41.6 & 71.0
& 77.2 & 46.3 & 87.6 \\
80\%  & 47.2 & 25.9 & 10.5 
& 71.7 & 44.5 & 79.0
& 77.4 & 47.1 & 89.7 \\
50\%  & 54.9 & 31.1 & 13.9
& 74.2 & 43.4 & 80.0
& 78.4 & 48.5 & 90.6 \\
20\%  & 63.0 & 34.6 & 26.9 
& 75.5 & 46.2 & 89.7
& 78.8 & 48.9 & 91.1 \\
0\%   & 68.2 & 37.0 & 40.3 
& 75.6 & 46.2 & 89.7
& 79.3 & 50.4 & 91.9 \\
\bottomrule
\end{tabular}
\end{table}

%% file: sec/discussion.tex
\section{Additional Discussions}



\textbf{Motivation for Studying Junk Data in Continual Pre-Training.}
Our study does not aim to simulate from-scratch pre-training on a naturally sampled web-scale corpus. Reliably estimating the proportion of junk or social-media text in modern LLM training mixtures is difficult because such mixtures are often proprietary and depend heavily on crawling scope, filtering criteria, deduplication, language selection, and sampling weights~\citep{penedo2023refinedweb}. Nevertheless, publicly documented pre-training corpora suggest that social and user-generated web text can constitute a non-trivial component of LLM-relevant data mixtures. For example, The Pile is an 825-GiB corpus composed of 22 subsets, in which social- and user-generated-content-style components account for approximately $17.24\%$ of the effective mixture by our calculation~\citep{gao2020pile}. This value is not intended as an estimate of proprietary LLM training corpora, but illustrates that user-generated web text can occupy a meaningful share of publicly documented pre-training mixtures.

Our focus is instead continual pre-training: we study whether further updating an already trained model on a controlled stream of junk-like web and social-media text can induce measurable behavioral changes. Accordingly, the junk-to-control ratios in our experiments represent intervention doses within the continual-pre-training corpus rather than estimates of the composition of the broader Internet. We match the token budgets of the junk and control data to isolate the effect of data type under the same update scale and vary the junk ratio to examine dose--response behavior.

Although the broader distribution of Internet text may dilute low-quality content during from-scratch pre-training, such dilution is less directly applicable to continual pre-training, which may use selected, newly collected, domain-specific, or source-specific data. Even if social-media junk constitutes a minority of all available web text, it may still exert a measurable effect when selected or upweighted in an update corpus. Twitter/X is particularly relevant in this context because it represents a high-volume source of user-generated text. Prior work reports hundreds of millions of active Twitter users and approximately 500 million tweets generated per day~\citep{antonakaki2021survey}. These statistics do not establish the exact fraction of Twitter/X content included in LLM training corpora, but motivate Twitter/X as a practically relevant source for studying the risks of junk-data continual pre-training.

\textbf{Twitter/X as an Initial Testbed.}
We select Twitter/X as the initial testbed for both practical and conceptual reasons. Twitter/X is a large-scale source of user-generated text~\citep{antonakaki2021survey} and is particularly suitable for studying short-form, engagement-driven social-media content. Its format naturally contains short posts, viral templates, attention-seeking language, and content shaped by rapid diffusion and algorithmically amplified engagement. The platform also provides observable engagement metadata, such as likes, reposts, replies, and quotes, which enables us to operationalize engagement-based junk content and construct controlled junk and control subsets.

These characteristics distinguish Twitter/X from platforms such as Reddit, where content is commonly organized around topic-specific communities and threaded discussions and may contain longer arguments, explanations, and exchanges of opinions. Twitter/X therefore provides a more direct setting for studying short, viral, and engagement-seeking text. This distinction does not imply that Reddit or other platforms are free from low-quality content; rather, different platform affordances may produce different forms of junk data.

Accordingly, our results should not be interpreted as evidence that all social-media platforms induce the same degree or type of degradation. The current study is a controlled investigation of short-form, engagement-driven Twitter/X text rather than a comprehensive cross-platform comparison. Other sources, including long-form discussion forums, short-video comment sections, question-answering communities, and heavily moderated platforms, may differ substantially in context length, repetition, moderation, toxicity, engagement incentives, and informational density. Cross-platform comparisons and mixtures of data from multiple platforms therefore constitute important directions for future work.

\textbf{Defining Engagement Metric M1.}
Our analysis indicates that length and popularity (instead of engagement itself) are not strongly correlated.
However, this is also a reason for us to include it as a combined metric for capturing \textbf{content-driven engagement}, which neither factor represents on its own: (1) Popularity reflects population-level engagement, which can arise from network effects or author influence rather than content quality. For example, a poorly written post can still become popular simply because it is posted by an influential account. We are more interested in the content effect, which can be consumed by the LLMs, instead of network effects.
(2) Shortness serves as a quantifiable proxy for semantic quality. As shown in Fig. 2, shorter texts tend to exhibit lower semantic richness. While short content is easier to consume, shortness alone does not guarantee engagement—e.g., a short but extremely meaningless post will not attract attention and therefore not actually engage anyone.

Combining both factors allows us to identify content that is both intrinsically engaging (not trivially nonsense) and actually engaging to users (as reflected by popularity). This aligns with our definition of content-driven engagement.
We note that M1 should not be interpreted as a semantics-free measure of engagement: although it is constructed from observable metadata, its length component may partially capture semantic richness or information density. The weak correlation between the two factors indicates that they capture distinct aspects of content-driven engagement. If they were highly correlated, combining them would provide limited additional information.



%% file: main.bib
@misc{oxford_brain_rot,
  author       = {{Oxford University Press}},
  title        = {{`Brain rot' named Oxford Word of the Year 2024}},
  year         = {2024},
  howpublished = {\url{https://corp.oup.com/news/brain-rot-named-oxford-word-of-the-year-2024/}}
}

@article{firth2019online_brain,
  title={The “online brain”: how the Internet may be changing our cognition},
  author={Firth, Joseph and Torous, John and Stubbs, Brendon and Firth, Josh A and Steiner, Genevieve Z and Smith, Lee and Alvarez-Jimenez, Mario and Gleeson, John and Vancampfort, Davy and Armitage, Christopher J and others},
  journal={World psychiatry},
  volume={18},
  number={2},
  pages={119--129},
  year={2019},
  publisher={Wiley Online Library}
}

@article{satici2023doomscrolling,
  title={Doomscrolling scale: Its association with personality traits, psychological distress, social media use, and wellbeing},
  author={Satici, Seydi Ahmet and Gocet Tekin, Emine and Deniz, M Engin and Satici, Begum},
  journal={Applied Research in Quality of Life},
  volume={18},
  number={2},
  pages={833--847},
  year={2023},
  publisher={Springer}
}

@article{hestness2017deep,
  title={Deep learning scaling is predictable, empirically},
  author={Hestness, Joel and Narang, Sharan and Ardalani, Newsha and Diamos, Gregory and Jun, Heewoo and Kianinejad, Hassan and Patwary, Md Mostofa Ali and Yang, Yang and Zhou, Yanqi},
  journal={arXiv preprint arXiv:1712.00409},
  year={2017}
}

@article{henighan2020scaling,
  title={Scaling laws for autoregressive generative modeling},
  author={Henighan, Tom and Kaplan, Jared and Katz, Mor and Chen, Mark and Hesse, Christopher and Jackson, Jacob and Jun, Heewoo and Brown, Tom B and Dhariwal, Prafulla and Gray, Scott and others},
  journal={arXiv preprint arXiv:2010.14701},
  year={2020}
}

@article{hoffmann2022training,
  title={Training compute-optimal large language models},
  author={Hoffmann, Jordan and Borgeaud, Sebastian and Mensch, Arthur and Buchatskaya, Elena and Cai, Trevor and Rutherford, Eliza and Casas, Diego de Las and Hendricks, Lisa Anne and Welbl, Johannes and Clark, Aidan and others},
  journal={arXiv preprint arXiv:2203.15556},
  year={2022}
}

@article{raghavendra2024revisiting,
  title={Revisiting the superficial alignment hypothesis},
  author={Raghavendra, Mohit and Nath, Vaskar and Hendryx, Sean},
  journal={arXiv preprint arXiv:2410.03717},
  year={2024}
}

@article{chen2025extracting,
  title={Extracting and understanding the superficial knowledge in alignment},
  author={Chen, Runjin and Perin, Gabriel Jacob and Chen, Xuxi and Chen, Xilun and Han, Yan and Hirata, Nina ST and Hong, Junyuan and Kailkhura, Bhavya},
  journal={arXiv preprint arXiv:2502.04602},
  year={2025}
}

@article{zhou2023lima,
  title={Lima: Less is more for alignment},
  author={Zhou, Chunting and Liu, Pengfei and Xu, Puxin and Iyer, Srinivasan and Sun, Jiao and Mao, Yuning and Ma, Xuezhe and Efrat, Avia and Yu, Ping and Yu, Lili and others},
  journal={Advances in Neural Information Processing Systems},
  volume={36},
  pages={55006--55021},
  year={2023}
}

@article{wang2025more,
  title={More is Less: The Pitfalls of Multi-Model Synthetic Preference Data in DPO Safety Alignment},
  author={Wang, Yifan and Chen, Runjin and Li, Bolian and Cho, David and Deng, Yihe and Zhang, Ruqi and Chen, Tianlong and Wang, Zhangyang and Grama, Ananth and Hong, Junyuan},
  journal={arXiv preprint arXiv:2504.02193},
  year={2025}
}

@inproceedings{
wettig2024qurating,
title={QuRating: Selecting High-Quality Data for Training Language Models},
author={Alexander Wettig and Aatmik Gupta and Saumya Malik and Danqi Chen},
booktitle={Forty-first International Conference on Machine Learning},
year={2024},
url={https://openreview.net/forum?id=GLGYYqPwjy}
}

@article{ra2018association,
  title={Association of digital media use with subsequent symptoms of attention-deficit/hyperactivity disorder among adolescents},
  author={Ra, Chaelin K and Cho, Junhan and Stone, Matthew D and De La Cerda, Julianne and Goldenson, Nicholas I and Moroney, Elizabeth and Tung, Irene and Lee, Steve S and Leventhal, Adam M},
  journal={Jama},
  volume={320},
  number={3},
  pages={255--263},
  year={2018},
  publisher={American Medical Association}
}

@article{sasaki2015anatomy,
  title={The anatomy of tweet overload: How number of tweets received, number of friends, and egocentric network density affect perceived information overload},
  author={Sasaki, Yuichi and Kawai, Daisuke and Kitamura, Satoshi},
  journal={Telematics and Informatics},
  volume={32},
  number={4},
  pages={853--861},
  year={2015},
  publisher={Elsevier}
}

@article{moisala2016media,
  title={Media multitasking is associated with distractibility and increased prefrontal activity in adolescents and young adults},
  author={Moisala, Mona and Salmela, Viljami and Hietaj{\"a}rvi, Lauri and Salo, Emma and Carlson, Synn{\"o}ve and Salonen, Oili and Lonka, Kirsti and Hakkarainen, Kai and Salmela-Aro, Katariina and Alho, Kimmo},
  journal={NeuroImage},
  volume={134},
  pages={113--121},
  year={2016},
  publisher={Elsevier}
}

@article{arc,
  title={Think you have solved question answering? try arc, the ai2 reasoning challenge},
  author={Clark, Peter and Cowhey, Isaac and Etzioni, Oren and Khot, Tushar and Sabharwal, Ashish and Schoenick, Carissa and Tafjord, Oyvind},
  journal={arXiv preprint arXiv:1803.05457},
  year={2018}
}

@article{ruler,
  title={RULER: What's the Real Context Size of Your Long-Context Language Models?},
  author={Hsieh, Cheng-Ping and Sun, Simeng and Kriman, Samuel and Acharya, Shantanu and Rekesh, Dima and Jia, Fei and Zhang, Yang and Ginsburg, Boris},
  journal={arXiv preprint arXiv:2404.06654},
  year={2024}
}

@article{hhrlhf,
  title={Training a helpful and harmless assistant with reinforcement learning from human feedback},
  author={Bai, Yuntao and Jones, Andy and Ndousse, Kamal and Askell, Amanda and Chen, Anna and DasSarma, Nova and Drain, Dawn and Fort, Stanislav and Ganguli, Deep and Henighan, Tom and others},
  journal={arXiv preprint arXiv:2204.05862},
  year={2022}
}

@article{advbench,
  title={Universal and transferable adversarial attacks on aligned language models},
  author={Zou, Andy and Wang, Zifan and Carlini, Nicholas and Nasr, Milad and Kolter, J Zico and Fredrikson, Matt},
  journal={arXiv preprint arXiv:2307.15043},
  year={2023}
}

@article{trait,
  title={Do llms have distinct and consistent personality? trait: Personality testset designed for llms with psychometrics},
  author={Lee, Seungbeen and Lim, Seungwon and Han, Seungju and Oh, Giyeong and Chae, Hyungjoo and Chung, Jiwan and Kim, Minju and Kwak, Beong-woo and Lee, Yeonsoo and Lee, Dongha and others},
  journal={arXiv preprint arXiv:2406.14703},
  year={2024}
}

@article{qi2023fine,
  title={Fine-tuning aligned language models compromises safety, even when users do not intend to!},
  author={Qi, Xiangyu and Zeng, Yi and Xie, Tinghao and Chen, Pin-Yu and Jia, Ruoxi and Mittal, Prateek and Henderson, Peter},
  journal={arXiv preprint arXiv:2310.03693},
  year={2023}
}

@article{yousef2025demystifying,
  title={Demystifying the new dilemma of brain rot in the digital era: A review},
  author={Yousef, Ahmed Mohamed Fahmy and Alshamy, Alsaeed and Tlili, Ahmed and Metwally, Ahmed Hosny Saleh},
  journal={Brain Sciences},
  volume={15},
  number={3},
  pages={283},
  year={2025},
  publisher={MDPI}
}

@article{vedechkina2021review,
  title={A review of evidence on the role of digital technology in shaping attention and cognitive control in children},
  author={Vedechkina, Maria and Borgonovi, Francesca},
  journal={Frontiers in psychology},
  volume={12},
  pages={611155},
  year={2021},
  publisher={Frontiers Media SA}
}

@article{haliti2024impact,
  title={Impact of Short Reels on Attention Span and Academic Performance of Undergraduate Students.},
  author={Haliti-Sylaj, Trendeline and Sadiku, Alisa},
  journal={Eurasian Journal of Applied Linguistics},
  volume={10},
  number={3},
  pages={60--68},
  year={2024},
  publisher={ERIC}
}

@article{saito2023verbosity,
  title={Verbosity bias in preference labeling by large language models},
  author={Saito, Keita and Wachi, Akifumi and Wataoka, Koki and Akimoto, Youhei},
  journal={arXiv preprint arXiv:2310.10076},
  year={2023}
}

@article{hu2024explaining,
  title={Explaining length bias in llm-based preference evaluations},
  author={Hu, Zhengyu and Song, Linxin and Zhang, Jieyu and Xiao, Zheyuan and Wang, Tianfu and Chen, Zhengyu and Yuan, Nicholas Jing and Lian, Jianxun and Ding, Kaize and Xiong, Hui},
  journal={arXiv preprint arXiv:2407.01085},
  year={2024}
}

@article{zheng2023judging,
  title={Judging llm-as-a-judge with mt-bench and chatbot arena},
  author={Zheng, Lianmin and Chiang, Wei-Lin and Sheng, Ying and Zhuang, Siyuan and Wu, Zhanghao and Zhuang, Yonghao and Lin, Zi and Li, Zhuohan and Li, Dacheng and Xing, Eric and others},
  journal={Advances in neural information processing systems},
  volume={36},
  pages={46595--46623},
  year={2023}
}

@article{wei2022chain,
  title={Chain-of-thought prompting elicits reasoning in large language models},
  author={Wei, Jason and Wang, Xuezhi and Schuurmans, Dale and Bosma, Maarten and Xia, Fei and Chi, Ed and Le, Quoc V and Zhou, Denny and others},
  journal={Advances in neural information processing systems},
  volume={35},
  pages={24824--24837},
  year={2022}
}

@misc{eliot2024brainrot,
  author       = {Lance Eliot},
  title        = {Generative AI and Brain Rot},
  year         = {2024},
  month        = jun,
  day          = {18},
  howpublished = {\url{https://www.forbes.com/sites/lanceeliot/2024/06/18/generative-ai-and-brain-rot/}},
  note         = {Forbes}
}

@article{binz2023using,
  title={Using cognitive psychology to understand GPT-3},
  author={Binz, Marcel and Schulz, Eric},
  journal={Proceedings of the National Academy of Sciences},
  volume={120},
  number={6},
  pages={e2218523120},
  year={2023},
  publisher={National Academy of Sciences}
}

@article{panda2024teach,
  title={Teach llms to phish: Stealing private information from language models},
  author={Panda, Ashwinee and Choquette-Choo, Christopher A and Zhang, Zhengming and Yang, Yaoqing and Mittal, Prateek},
  journal={arXiv preprint arXiv:2403.00871},
  year={2024}
}

@article{shumailov2023curse,
  title={The curse of recursion: Training on generated data makes models forget},
  author={Shumailov, Ilia and Shumaylov, Zakhar and Zhao, Yiren and Gal, Yarin and Papernot, Nicolas and Anderson, Ross},
  journal={arXiv preprint arXiv:2305.17493},
  year={2023}
}

@article{shumailov2024ai,
  title={AI models collapse when trained on recursively generated data},
  author={Shumailov, Ilia and Shumaylov, Zakhar and Zhao, Yiren and Papernot, Nicolas and Anderson, Ross and Gal, Yarin},
  journal={Nature},
  volume={631},
  number={8022},
  pages={755--759},
  year={2024},
  publisher={Nature Publishing Group UK London}
}

@article{seddik2024bad,
  title={How bad is training on synthetic data? a statistical analysis of language model collapse},
  author={Seddik, Mohamed El Amine and Chen, Suei-Wen and Hayou, Soufiane and Youssef, Pierre and Debbah, Merouane},
  journal={arXiv preprint arXiv:2404.05090},
  year={2024}
}

@article{zhang2024persistent,
  title={Persistent pre-training poisoning of LLMs},
  author={Zhang, Yiming and Rando, Javier and Evtimov, Ivan and Chi, Jianfeng and Smith, Eric Michael and Carlini, Nicholas and Tram{\`e}r, Florian and Ippolito, Daphne},
  journal={arXiv preprint arXiv:2410.13722},
  year={2024}
}

@article{fu2024poisonbench,
  title={Poisonbench: Assessing large language model vulnerability to data poisoning},
  author={Fu, Tingchen and Sharma, Mrinank and Torr, Philip and Cohen, Shay B and Krueger, David and Barez, Fazl},
  journal={arXiv preprint arXiv:2410.08811},
  year={2024}
}

@article{bai2022constitutional,
  title={Constitutional ai: Harmlessness from ai feedback},
  author={Bai, Yuntao and Kadavath, Saurav and Kundu, Sandipan and Askell, Amanda and Kernion, Jackson and Jones, Andy and Chen, Anna and Goldie, Anna and Mirhoseini, Azalia and McKinnon, Cameron and others},
  journal={arXiv preprint arXiv:2212.08073},
  year={2022}
}

@article{christiano2017deep,
  title={Deep reinforcement learning from human preferences},
  author={Christiano, Paul F and Leike, Jan and Brown, Tom and Martic, Miljan and Legg, Shane and Amodei, Dario},
  journal={Advances in neural information processing systems},
  volume={30},
  year={2017}
}

@article{milli2025engagement,
  title={Engagement, user satisfaction, and the amplification of divisive content on social media},
  author={Milli, Smitha and Carroll, Micah and Wang, Yike and Pandey, Sashrika and Zhao, Sebastian and Dragan, Anca D},
  journal={PNAS nexus},
  volume={4},
  number={3},
  pages={pgaf062},
  year={2025},
  publisher={Oxford University Press US}
}

@inproceedings{suh2010want,
  title={Want to be retweeted? large scale analytics on factors impacting retweet in twitter network},
  author={Suh, Bongwon and Hong, Lichan and Pirolli, Peter and Chi, Ed H},
  booktitle={2010 IEEE second international conference on social computing},
  pages={177--184},
  year={2010},
  organization={IEEE}
}

@misc{twitter2023recommendation,
  author       = {{X Corp.}},
  title        = {Twitter's Recommendation Algorithm},
  year         = {2023},
  month        = mar,
  howpublished = {\url{https://blog.x.com/engineering/en_us/topics/open-source/2023/twitter-recommendation-algorithm}},
  note         = {Accessed: 2025-09-20}
}

@article{malhotra2011get,
  title={How to get your messages retweeted},
  author={Malhotra, Arvind and Malhotra, Claudia Kubowicz and See, Alan},
  journal={MIT Sloan Management Review},
  year={2011}
}

@article{llama3,
  title={The llama 3 herd of models},
  author={Grattafiori, Aaron and Dubey, Abhimanyu and Jauhri, Abhinav and Pandey, Abhinav and Kadian, Abhishek and Al-Dahle, Ahmad and Letman, Aiesha and Mathur, Akhil and Schelten, Alan and Vaughan, Alex and others},
  journal={arXiv preprint arXiv:2407.21783},
  year={2024}
}

@article{qwen25,
      title={Qwen2.5 Technical Report}, 
      author={Qwen and : and An Yang and Baosong Yang and Beichen Zhang and Binyuan Hui and Bo Zheng and Bowen Yu and Chengyuan Li and Dayiheng Liu and Fei Huang and Haoran Wei and Huan Lin and Jian Yang and Jianhong Tu and Jianwei Zhang and Jianxin Yang and Jiaxi Yang and Jingren Zhou and Junyang Lin and Kai Dang and Keming Lu and Keqin Bao and Kexin Yang and Le Yu and Mei Li and Mingfeng Xue and Pei Zhang and Qin Zhu and Rui Men and Runji Lin and Tianhao Li and Tianyi Tang and Tingyu Xia and Xingzhang Ren and Xuancheng Ren and Yang Fan and Yang Su and Yichang Zhang and Yu Wan and Yuqiong Liu and Zeyu Cui and Zhenru Zhang and Zihan Qiu},
      year={2025},
      journal={arXiv preprint arXiv:2412.15115},
}

@article{qwen3,
  title={Qwen3 technical report},
  author={An Yang and Anfeng Li and Baosong Yang and Beichen Zhang and Binyuan Hui and Bo Zheng and Bowen Yu and Chang Gao and Chengen Huang and Chenxu Lv and Chujie Zheng and Dayiheng Liu and Fan Zhou and Fei Huang and Feng Hu and Hao Ge and Haoran Wei and Huan Lin and Jialong Tang and Jian Yang and Jianhong Tu and Jianwei Zhang and Jianxin Yang and Jiaxi Yang and Jing Zhou and Jingren Zhou and Junyang Lin and Kai Dang and Keqin Bao and Kexin Yang and Le Yu and Lianghao Deng and Mei Li and Mingfeng Xue and Mingze Li and Pei Zhang and Peng Wang and Qin Zhu and Rui Men and Ruize Gao and Shixuan Liu and Shuang Luo and Tianhao Li and Tianyi Tang and Wenbiao Yin and Xingzhang Ren and Xinyu Wang and Xinyu Zhang and Xuancheng Ren and Yang Fan and Yang Su and Yichang Zhang and Yinger Zhang and Yu Wan and Yuqiong Liu and Zekun Wang and Zeyu Cui and Zhenru Zhang and Zhipeng Zhou and Zihan Qiu},
  journal={arXiv preprint arXiv:2505.09388},
  year={2025}
}

@article{pile,
    title={The {P}ile: An 800GB Dataset of Diverse Text for Language Modeling},
    author={Gao, Leo and Biderman, Stella and Black, Sid and Golding, Laurence and Hoppe, Travis and Foster, Charles and Phang, Jason and He, Horace and Thite, Anish and Nabeshima, Noa and Presser, Shawn and Leahy, Connor},
    journal={arXiv preprint arXiv:2101.00027},
    year={2020}
}

@article{shinn2023reflexion,
  title={Reflexion: Language agents with verbal reinforcement learning},
  author={Shinn, Noah and Cassano, Federico and Gopinath, Ashwin and Narasimhan, Karthik and Yao, Shunyu},
  journal={Advances in Neural Information Processing Systems},
  volume={36},
  pages={8634--8652},
  year={2023}
}

@inproceedings{ye2025auditing,
  title={Auditing Political Exposure Bias: Algorithmic Amplification on Twitter/X During the 2024 US Presidential Election},
  author={Ye, Jinyi and Luceri, Luca and Ferrara, Emilio},
  booktitle={Proceedings of the 2025 ACM Conference on Fairness, Accountability, and Transparency},
  pages={2349--2362},
  year={2025}
}

@article{khattab2023dspy,
  title={Dspy: Compiling declarative language model calls into self-improving pipelines},
  author={Khattab, Omar and Singhvi, Arnav and Maheshwari, Paridhi and Zhang, Zhiyuan and Santhanam, Keshav and Vardhamanan, Sri and Haq, Saiful and Sharma, Ashutosh and Joshi, Thomas T and Moazam, Hanna and others},
  journal={arXiv preprint arXiv:2310.03714},
  year={2023}
}

@misc{alpaca,
  author = {Rohan Taori and Ishaan Gulrajani and Tianyi Zhang and Yann Dubois and Xuechen Li and Carlos Guestrin and Percy Liang and Tatsunori B. Hashimoto },
  title = {Stanford Alpaca: An Instruction-following LLaMA model},
  year = {2023},
  publisher = {GitHub},
  journal = {GitHub repository},
  howpublished = {\url{https://github.com/tatsu-lab/stanford_alpaca}},
}

@article{zheng2025lifelong,
author = {Zheng, Junhao and Qiu, Shengjie and Shi, Chengming and Ma, Qianli},
title = {Towards Lifelong Learning of Large Language Models: A Survey},
year = {2025},
issue_date = {August 2025},
publisher = {ACM},
journal = {ACM Comput. Surv.},
month = mar,
articleno = {193},
numpages = {35},
}

@article{c4,
  title={Exploring the limits of transfer learning with a unified text-to-text transformer},
  author={Raffel, Colin and Shazeer, Noam and Roberts, Adam and Lee, Katherine and Narang, Sharan and Matena, Michael and Zhou, Yanqi and Li, Wei and Liu, Peter J},
  journal={Journal of machine learning research},
  volume={21},
  number={140},
  pages={1--67},
  year={2020}
}

@article{penedo2023refinedweb,
  title={The refinedweb dataset for falcon llm: Outperforming curated corpora with web data only},
  author={Penedo, Guilherme and Malartic, Quentin and Hesslow, Daniel and Cojocaru, Ruxandra and Alobeidli, Hamza and Cappelli, Alessandro and Pannier, Baptiste and Almazrouei, Ebtesam and Launay, Julien},
  journal={Advances in Neural Information Processing Systems},
  volume={36},
  pages={79155--79172},
  year={2023}
}

@article{gao2020pile,
  title={The pile: An 800gb dataset of diverse text for language modeling},
  author={Gao, Leo and Biderman, Stella and Black, Sid and Golding, Laurence and Hoppe, Travis and Foster, Charles and Phang, Jason and He, Horace and Thite, Anish and Nabeshima, Noa and others},
  journal={arXiv preprint arXiv:2101.00027},
  year={2020}
}

@article{antonakaki2021survey,
  title={A survey of Twitter research: Data model, graph structure, sentiment analysis and attacks},
  author={Antonakaki, Despoina and Fragopoulou, Paraskevi and Ioannidis, Sotiris},
  journal={Expert systems with applications},
  volume={164},
  pages={114006},
  year={2021},
  publisher={Elsevier}
}
